%% file: main.tex
\pdfoutput=1

\documentclass[11pt]{article}

\usepackage[preprint]{acl}

\usepackage{times}
\usepackage{latexsym}

\usepackage[T1]{fontenc}

\usepackage[utf8]{inputenc}

\usepackage{microtype}

\usepackage{inconsolata}

\usepackage{microtype}
\usepackage{graphicx}
\usepackage{subfigure}
\usepackage{booktabs}
\usepackage{amsmath}
\usepackage{amssymb}
\usepackage{mathtools}
\usepackage{amsthm}
\usepackage{amsfonts}
\usepackage{multirow}
\usepackage[frozencache=true, cachedir=minted-cache]{minted}
\usepackage{enumitem}
\input{custom}

\input{math_commands.tex}

\newenvironment{lmttfont}{\fontfamily{lmtt}\selectfont}{\par}

\title{Learning to Generate Instruction Tuning Datasets for\\
Zero-Shot Task Adaptation}

\author{Nihal V. Nayak\;\;\;Yiyang Nan\;\;\;Avi Trost\;\;\;Stephen H. Bach\\
  Department of Computer Science, Brown University\\
  \texttt{\{nnayak2, ynan3, atrost, sbach\}@cs.brown.edu}
}

\begin{document}
\maketitle

\input{sections/abstract}

\input{sections/introduction}

\input{sections/zsta}
\input{sections/related_work}
\input{sections/method}

\input{sections/experiments}
\input{sections/analysis}
\input{sections/additional}

\input{sections/conclusion}

\section*{Acknowledgements}
We appreciate Yeganeh Kordi, Zheng-Xin Yong, Aidan LaBella, and members of the Brown SuperLab for their thoughtful comments and feedback on our draft. 
We gratefully acknowledge support from Cisco. Disclosure: Stephen
Bach is an advisor to Snorkel AI, a company that provides software and services for data-centric
artificial intelligence.

\section*{Limitations}
Our work relies on the availability of large amounts of unannotated text. 
If only a small quantity of unannotated text is present, the target language model, after adaptation, may experience a drop in performance. 
While we demonstrate positive improvements on pretrained and instruction-tuned models, our observations are limited to the three task types considered in our experiments.

\section*{Potential Risks}
\sys poses risks similar to those of any large language model.
For example, our model could be used to generate factually incorrect datasets in specialized domains.
Our model can exhibit the biases and stereotypes of the base model, Mistral-7B, even after extensive supervised fine-tuning.
Finally, our model does not include safety training and can potentially generate harmful content.

\def\UrlBreaks{\do\/\do-\do\&\do.\do:}
\bibliography{main}
\clearpage
\appendix
\input{appendix/appendix}

\end{document}

%% file: custom.tex
\usepackage{xspace}
\newcommand{\sys}{Bonito\xspace}
\definecolor{increase_color}{HTML}{27ae60}
\definecolor{decrease_color}{HTML}{0984e3}
\newcommand{\increase}[2][increase_color]{{\textcolor{#1}{\textbf{+#2}}}}
\newcommand{\decrease}[2][red]{{\textcolor{#1}{\textbf{-#2}}}}

\definecolor{bg}{rgb}{0.97,0.97,0.97}
\definecolor{bga}{rgb}{0.97,0.97,0.97}
\definecolor{bgb}{rgb}{0.97,0.97,0.97}

%% file: math_commands.tex
\usepackage{amsmath,amsfonts,bm}

\def\eqref#1{equation~\ref{#1}}

\def\1{\bm{1}}

\DeclareMathAlphabet{\mathsfit}{\encodingdefault}{\sfdefault}{m}{sl}
\SetMathAlphabet{\mathsfit}{bold}{\encodingdefault}{\sfdefault}{bx}{n}

%% file: sections/abstract.tex
\begin{abstract}
We introduce Bonito, an open-source model for \emph{conditional task generation} that converts unannotated text into task-specific training datasets for instruction tuning. 
We aim to enable zero-shot task adaptation of large language models on users' specialized, private data.
We train Bonito by fine-tuning a pretrained large language model on a new large-scale dataset with 1.65M examples created by remixing existing instruction tuning datasets into \emph{meta-templates}.
The meta-templates for a dataset produce training examples where the input is the unannotated text and the task attribute and the output consists of the instruction and the response.
We use Bonito to generate synthetic tasks for seven datasets from specialized domains with unannotated text across three task types---yes-no question answering, extractive question answering, and natural language inference---and adapt language models. 
We show that Bonito significantly improves the average performance of pretrained and instruction tuned models over the de facto self supervised baseline.  
For example, adapting Mistral-Instruct-v2 and instruction tuned variants of Mistral and Llama2 with Bonito improves the strong zero-shot performance by 22.1 F1 points whereas the next word prediction objective undoes some of the benefits of instruction tuning and reduces the average performance by 0.8 F1 points.
We conduct additional experiments with Bonito to understand the effects of the domain, the size of the training set, and the choice of alternative synthetic task generators. 
Overall, we show that learning with synthetic instruction tuning datasets is an effective way to adapt language models to new domains.
The model, dataset, and code are available at \href{https://github.com/BatsResearch/bonito}{https://github.com/BatsResearch/bonito}.
\end{abstract}

%% file: sections/introduction.tex
\input{fig/workflow}
\section{Introduction}
Large language models show remarkable zero-shot capabilities by simply learning to predict the next token at scale~\citep{brown:arxiv20,touvron:arxiv23}.
By fine-tuning these models on instruction tuning datasets containing many \emph{tasks}---each comprising an input \emph{instruction} and a desired \emph{response}---the model generally improves in its ability to respond to unseen instructions.
However, this generalization is still limited by the qualities of the instruction tuning dataset.
Existing datasets like Public Pool of Prompts (P3)~\citep{bach:acldemo22}, Natural Instructions~\citep{mishra:arxiv21,wang:emnlp22}, and Dolly-v2~\citep{conover:arxiv23} focus on text from the Web and classic natural language tasks so that they can serve a wide range of use cases, i.e., they are a one-size-fits-all approach. 
On the other hand, tasks in areas like biomedical and legal domains require specialized, often implicit, domain knowledge.
We study how to adapt language models to follow instructions in specialized domains without annotated data. 

The ability to follow task-specific instructions in specialized domains is important for bringing the benefits of large language models to a wider range of users.
Recent evaluations---including evaluations of proprietary models---show that they often significantly underperform specialized models~\citep{kocon:infofusion23,shen:arxiv23,ziems:arxiv23}, particularly in domains requiring subject matter expertise.
This motivates us to investigate effective ways to provide domain knowledge to large language models. 

Self supervision in the form of next word prediction on the target corpus is a simple way to teach language models about new domains~\citep{gururangan:acl20}.
However, this approach requires an enormous amount of training to achieve strong performance~\citep{chen:arxiv23}. 
Further, we find that self supervision can undo the benefits of instruction tuning (see Section \ref{sec:experiments:instruction_tuned}). 
Alternatively, continued training of models with instructions from specialized domains significantly improves performance ~\citep{scialom:emnlp22,shi:neurips23,yunxiang:arxiv23,deng:arxiv23,singhal:nature23,wu:arxiv24}. 
However, they need to repeat the time-consuming and labor-intensive process of annotating a domain-specific dataset.
Furthermore, collecting instructions in specialized domains is very expensive because they are annotated by domain experts such as scientists and researchers~\citep{thulke:arxiv24}. 
In this work, we automate the creation of instruction tuning datasets in specialized domains.

We create \sys, an open-source model to convert unannotated text from specialized domains into task-specific training datasets for instruction tuning (Figure~\ref{fig:overview}). 
We call this problem \emph{conditional task generation}.
Our key idea is to make a new large-scale dataset called Conditional Task Generation with Attributes (CTGA), to train Bonito, by reorganizing existing instruction tuning datasets (see Figure \ref{fig:ctga_construction}). 
Instruction tuning datasets like P3~\citep{bach:acldemo22} exist as templates that convert semi-structured examples of natural language tasks into a fully prompted format, in which both the input and the desired response are text strings.
We focus on a subset of the templates in P3 that require a \emph{context} or a passage to complete the task. 
For example, a context could be a paragraph that contains a fact or that contains the answer to a question.
Then, we remix these templates to create the meta-templates.
Each meta-template for a dataset produces training examples in which the input is context and a task attribute such as yes-no question answering, and the output is the entire task: the instruction (including the context) and the desired response.
In this way, we can easily create abundant, diverse examples to train Bonito. 
After training Bonito, we can use new unannotated text from the target domain as the context to generate task-specific synthetic datasets and train specialized language models.

\sys significantly improves over self supervision on zero-shot task adaptation of pretrained and instruction tuned models.  
We use \sys to generate instruction tuning data for seven datasets across three task types---yes-no question answering (PubMedQA and Privacy Policy QA), extractive question answering (SQuADShifts-NYT, Amazon, and Reddit), and natural language inference (ContractNLI and Vitamin C)---and adapt language models. 
Our results show that \sys improved Mistral-7B by 34.7 F1 points and Llama 2 7B by 31.6 F1 points over the self supervised baseline, next word prediction objective. 
We also consider a more practical setting where we further train Mistral-7B-Instruct-v0.2 and instruction tuned variants of Mistral-7B and Llama 2 7B trained on the T0 split of the P3 dataset. 
Our results show that \sys outperforms the strong zero-shot baseline performance by an average of 22.1 F1 points across all the models. 
On the other hand, we find that self supervision undoes some of the benefits of instruction tuning, i.e., it leads to catastrophic forgetting, resulting in a drop in performance by an average of 0.8 F1 points across all models. 
Our analysis of Bonito shows that even task specialized models can be further improved by simply learning on \sys generated tasks (see Section \ref{sec:analysis:self_distillation}).  
We also find that training with more synthetic instructions on datasets like PubMedQA and Vitamin C improves model performance the most compared to other datasets (see Section \ref{sec:analysis:training_size}). 
We perform additional experiments by prompting off-the-shelf open-source models like Zephyr-7B-$\beta$ and Mistral-7B-Instruct-v0.2 and GPT-4 to generate tasks and find they can often improve the pretrained models but still struggle to increase model performance further when they are instruction tuned (see Section \ref{sec:additional}). 
Finally, our human evaluation of Bonito-generated tasks shows that Bonito and GPT-4o generate the same answer on 71 to 77 percent of the tasks. 
This indicates that Bonito generates high-quality tasks. 
However, there is room for improvement in generation quality to increase downstream model performance.

In summary, our main contributions are: 
\begin{itemize}    
    \item We introduce \sys, an open-source model for conditional task generation model to convert the user's unannotated text into task-specific instruction tuning datasets.
    
    \item Our experiments on zero-shot task adaptation on seven datasets across three task types show that \sys improves over the self supervised baseline by an average of 33.1 F1 points on the pretrained models and 22.9 F1 points on the instruction tuned models. 
    
    \item We analyze the effect of the domain, training size, and the choice of alternative task generators highlighting the benefits and limitations of \sys.

\end{itemize}

%% file: fig/workflow.tex
\begin{figure*}[t]
    \centering
    \includegraphics[width=1\textwidth]{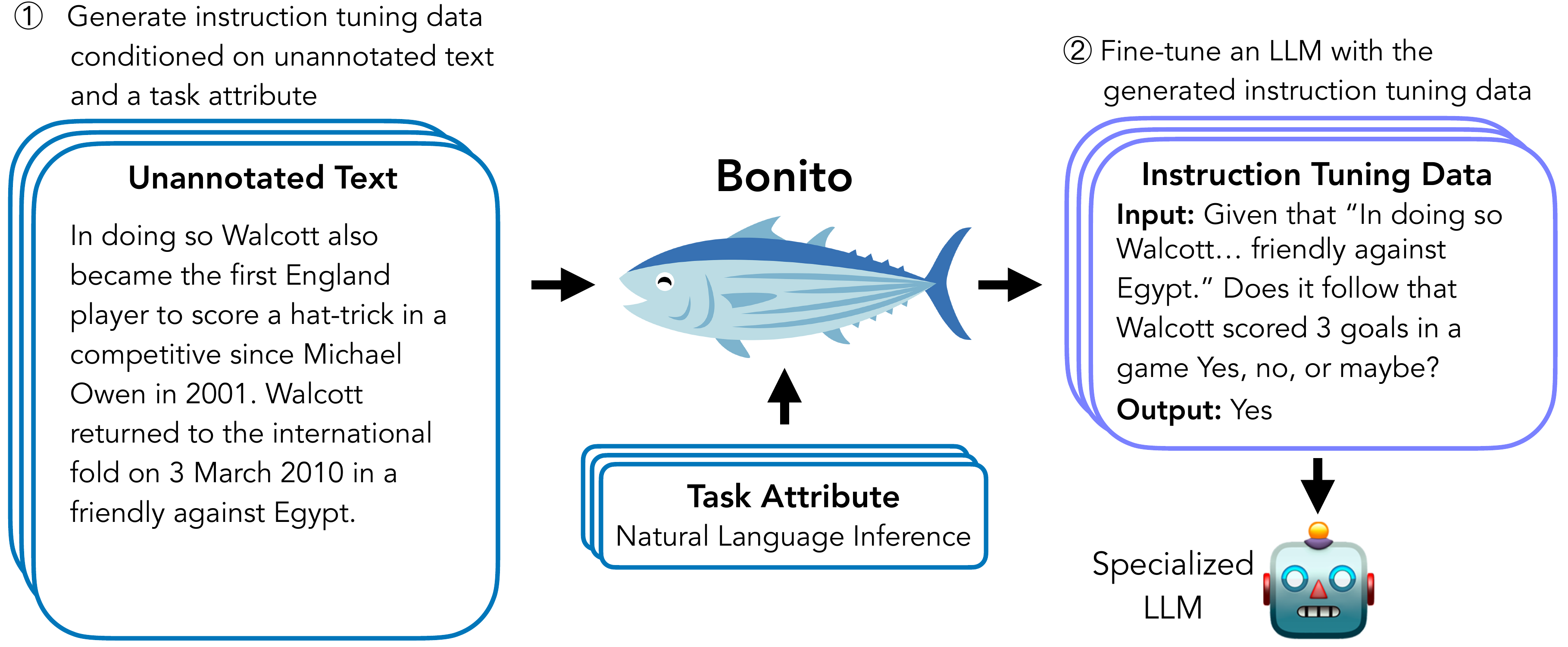}
    \caption{
    Bonito workflow for conditional task generation and adaptation.
    Bonito takes unannotated text as input, along with task attributes, to generate instruction tuning data.
    For each unannotated text, it generates an instruction that references the text and a target response.
    The instruction tuning data is then used to (further) fine-tune a language model, adapting it to the task in the specialized domain.
    }
    \label{fig:overview}
\end{figure*}

%% file: sections/zsta.tex
\section{Zero-Shot Task Adaptation}
We describe the problem of zero-shot task adaptation. 
We have a language model, either pretrained via self supervision or further fine-tuned on a training mixture like P3~\citep{bach:acldemo22}, along with a corpus of unannotated text from the target domain.
We also know the target task type e.g., extractive question answering, natural language inference, etc.
If the target task type has a fixed set of labels such as ``yes'' or ``no'' in yes-no question answering, we assume access to the label space.  
Our goal is to adapt the language model to follow task instructions in the target domain without human annotations, achieving zero-shot task adaptation.

%% file: sections/related_work.tex
\section{Related Work}

\paragraph{Instruction Tuning}
Multitask instruction tuning of language models dramatically improves their ability to follow instructions and generalize to new unseen tasks \citep{sanh:iclr22,wei:iclr22,mishra:arxiv21,longpre:icml23,chung:arxiv22,zhou:arxiv23,li:arxiv23}.
Typically, pretrained models are trained to follow instructions on large-scale training mixtures such as P3 \citep{bach:acldemo22} and the FLAN collection~\citep{longpre:icml23}. 
In this work, we use P3 to create meta-templates and train \sys to generate NLP tasks in specialized domains.

\paragraph{Domain Adaptation}
Several works have adapted large language models to tasks in specialized domains \citep{gururangan:acl20,yunxiang:arxiv23,cui:arxiv23,wu:arxiv23}.
Several works~\citep{gu:acm22,chen:arxiv23} show that self supervision or continuing the pretraining objective of the pretrained language model on the target domain corpus improves downstream performance.
In this work, we find that self supervision improves the performance of pretrained models but hurts the performance of instruction tuned models (Section \ref{sec:experiments}). 

Recent work has adapted language models by training on large-scale in-domain datasets\citep{parmar:naacl22,gupta:emnlp22,singhal:arxiv23,deng:arxiv23} or with a few examples from domain-specific tasks~\citep{singhal:nature23}.
However, annotating training datasets for new domains is labor-intensive and expensive~\citep{thulke:arxiv24}. 
We focus on generating training datasets in specialized domains and adapting language models without annotations. 

Zero-shot task adaptation is closely related to unsupervised domain adaptation~\citep{ganin:icml15}.
In unsupervised domain adaptation, a trained model is used to generate pseudo-labels for the target unlabeled data and then trained on these labels.
Naive pseudo-labeling cannot be applied to this work since we consider tasks like question answering and natural language inference tasks that require a question or a hypothesis before producing an answer in natural language. 
Further, popular techniques used in unsupervised domain adaptation such as choosing top-K confident classes~\citep{huang:arxiv22,menghini:neurips23} cannot be easily adapted to NLP tasks as there may not be an explicit notion of classes. 

There is a growing interest in using retrieval augmented generation (RAG) for domain-specific question answering~\citep{lewis:neurips20,karpukhin:emnlp20,siriwardhana:tacl23,zhang:arxiv24}. 
In a RAG pipeline, given a question, the most relevant documents are retrieved before accurately producing an answer with a language model. 
Our work compliments the RAG pipeline as we assume access to the gold documents or paragraphs from specialized domains and improve the language model's ability to answer the questions. 

\paragraph{Task Generation}
Task generation is a fast-growing area of research to adapt large language models to follow instructions \citep{wang:acl23,alpaca:github22,honovich:acl23,koksal:arxiv23,kang:arxiv23,liu:arxiv24}.
They typically condition a model on a set of seed task demonstrations and generate new synthetic tasks \citep{wang:acl23,honovich:acl23,kang:arxiv23}.
However, task generation conditioned on the user's unannotated text has mostly been ignored. 
Additionally, generating tasks with API-based models like GPT is expensive and cannot be used for proprietary or private research data~\citep{koksal:arxiv23}.
On the other hand, \sys is an open-source model that can be used to create tasks with the user's unannotated text without additional API costs.

Recently, \citet{li:arxiv23} proposed to learn a backtranslation model, similar to \sys, to iteratively grow and refine their instruction tuning dataset~\citep{gulcehre:arxiv23}. 
However, they focus on generating instructions conditioned on the unannotated text from a web corpus for long-form conversational data where the answer to the instruction is the unannotated text. 
In contrast, we focus on generating NLP tasks conditioned on a task type and unannotated text from a specialized domain.
Further, we consider tasks such as question answering and natural language inference that require a question or a hypothesis before generating the appropriate answer. 
\input{fig/training_example}

Concurrent to this work,~\citet{yehudai:iclr24} use in-context learning with Falcon-40B and Llama-65B to generate ``grounded tasks'' to adapt smaller models like FLAN-T5-XL (3B).
These grounded tasks are similar to conditional tasks, except the instructions do not necessarily refer directly to the user's text.
They might only be based on it, such as asking an open-ended question based on the original text.
Our work goes further in several ways.
First, we study how to create an open-source model for conditional task generation, as opposed to relying on prompting alone.
Second, Bonito has only 7B parameters and we show that it creates data that can improve instruction tuned models of the same size and outperform even larger models like Flan-T5-XXL (11B) (see Appendix \ref{app:flan}).
Third, we evaluate tasks with precise correct/incorrect answers, such as yes-no question answering and natural language inference, as opposed to tasks evaluated with similarity metrics.

\paragraph{Knowledge Distillation}
Knowledge distillation is a well-studied area~\citep{hinton:arxiv15,sanh:arxiv19,he:arxiv19}. 
Typically, smaller models learn from the outputs of a larger model.
Most recently, API-based models have been used to generate tasks and distilled into smaller models to mimic the abilities of the API-based models \citep{peng:arxiv23,gudibande:arxiv23}.
In this work, we use \sys to generate tasks based on the user's context and distill them into pretrained and instruction tuned models of the same size for zero-shot task adaptation (see Section \ref{sec:experiments}).

\paragraph{Question Generation}
Several works have been proposed in question generation over the years \citep{ha:naacl03,pan:acl20,lewis:tacl21,ushio:acl23}.
They often use heuristics such as templates~\citep{ha:naacl03}, named entity recognition ~\cite{lewis:tacl21},
and semantic graphs~\citep{pan:acl20}.
In our work, we train a language model without relying on task-specific heuristics. 
\citet{ushio:acl23} is closely related to our work as they train a unified model to generate extractive questions and answers, but only focus on adapting small pretrained language models like T5-Large (770M).
In contrast, \sys can generate tasks beyond extractive question answering and enable zero-shot task adaptation on several task types with large models like Llama 2 7B and Mistral-7B. 

%% file: fig/training_example.tex
\begin{figure*}[t]
    \centering
    \includegraphics[width=1.0\linewidth]{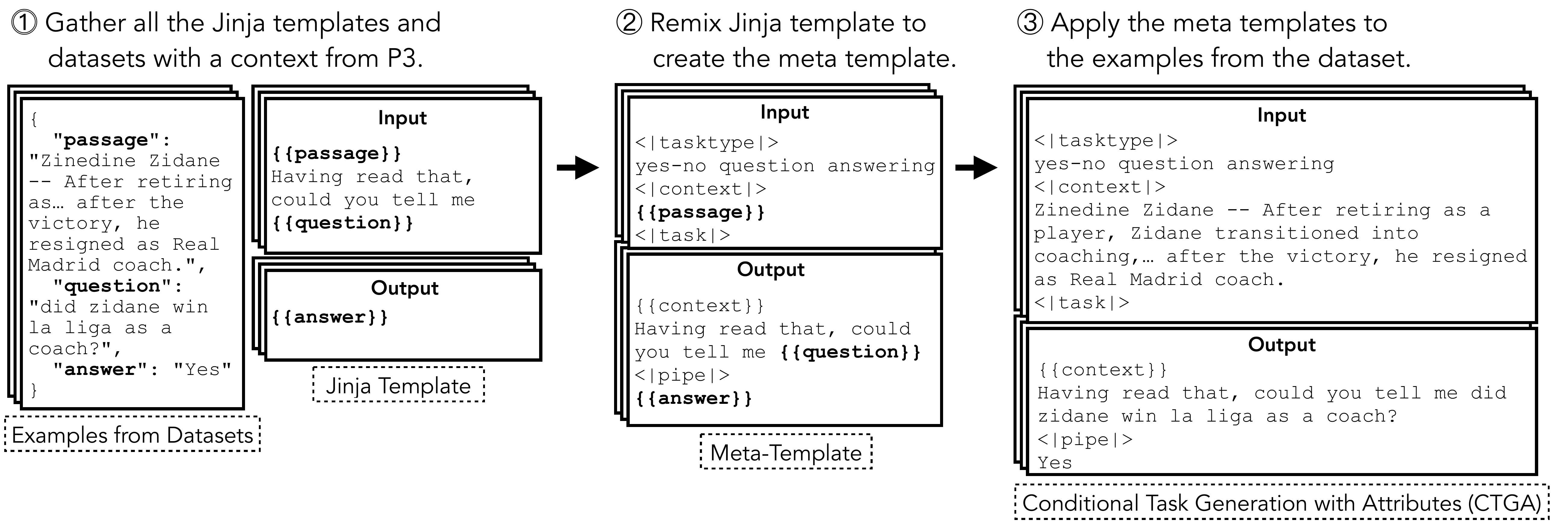}
    \caption{The high-level process of constructing the Conditional Task Generation with Attributes (CTGA) dataset. 
    }
    \label{fig:ctga_construction}
\end{figure*}

%% file: sections/method.tex
\section{\sys: Learning to Generate Tasks}
We describe the steps to create the Conditional Task Generation with Attributes (CTGA) dataset and train \sys. 
Then, we briefly describe the procedure to create synthetic instruction tuning datasets with the target unannotated texts to adapt language models.

\paragraph{Key Properties}
We outline the key properties that we desire in our conditional task generation model: (1) given a corpus containing articles and paragraphs, the model should take the text as input and generate high-quality tasks that require minimal cleaning or post-processing, (2) the model should adhere to the task type like extractive question answering or natural language inference task, and (3) the model should generate diverse tasks for the exact text with varying styles. 

\subsection{Creating Bonito: Dataset and Training}

To create a model that generates tasks conditioned on text, we create a new training dataset: Conditional Task Generation with Attributes (CTGA).
The dataset contains 1.65 million examples derived from P3~\citep{bach:acldemo22} by annotating 323 prompt templates from 38 datasets with 16 task types (see Appendix \ref{app:ctga}). 
Then, we train a pretrained large language model on this training dataset to create Bonito.

\paragraph{Constructing the Dataset}
Figure \ref{fig:ctga_construction} shows the process of constructing the Conditional Task Generation with Attributes (CTGA) dataset. 
First, we identify datasets from P3~\citep{bach:acldemo22} that require a passage or a context to complete the task. 
For example, SQuAD \citep{rajpurkar:emnlp2016} requires a context to answer extractive question answering tasks, whereas CommonSenseQA \citep{talmor:acl19} asks a multiple choice question without providing any relevant text. 
We identify a total of 38 datasets to be included in CTGA. 
For each dataset, we also collect the Jinja\footnote{\href{https://jinja.palletsprojects.com/en/3.1.x/}{https://jinja.palletsprojects.com/en/3.1.x/}} templates from P3.
Next, we remix the Jinja templates to create meta-templates. 
A meta-template is a Jinja template that includes the task attribute or the task type and the key for the context column in the input and the Jinja template for the instruction-response pair in the output with a placeholder \texttt{\{\{context\}\}} to avoid repeating the context. 
Since Jinja templates from P3 do not include a task type, we manually annotate them with a target task type such as yes-no question answering (see Appendix \ref{app:ctga} for details).
Overall, we get 323 meta-templates spanning 16 task types (See Table \ref{tab:task_distribution} for the list of task types). 
Finally, we apply the meta-templates to all the examples in a dataset to create the CTGA dataset, i.e., we replace the keys for the columns in the Jinja templates with corresponding key-value pairs from the examples.
If the dataset has multiple meta-templates, we uniformly sample one meta-template per example. 
We limit the total number of examples per dataset to 100,000. 
The final training dataset is used to train \sys.

\paragraph{Training the \sys Model}\label{sec:bonito:training}
We train \sys by fine-tuning Mistral-7B, an open-source decoder language model~\citep{jiang:arxiv23}, on the CTGA dataset.
The model is trained by optimizing the cross entropy loss over the output tokens.
We include all the hyperparameters and training details in Appendix \ref{app:hyp:bonito}.

\subsection{Adapting Models with Bonito}
We use Bonito to create synthetic instruction tuning datasets for the target unannotated texts. 
Then, the target language model is adapted by training on the synthetic dataset to get the specialized language model.

\paragraph{Generating the Synthetic Dataset}
Figure \ref{fig:overview} shows the inference with Bonito to generate the synthetic instruction tuning dataset.
The unannotated text and the task type are passed to the Bonito model to get the synthetic instruction-response pairs. 
The process is repeated for all the unannotated text to get the training dataset. 
The generated pairs are then post-processed into a standardized instruction-response format for instruction tuning.
In each generation, we replace \texttt{\{\{context\}\}} with the corresponding unannotated text from the input.
If the generated output is not parsable due to missing \texttt{<|pipe|>}, we filter them out.

\paragraph{Adapting the Target Model} 
We train the target language model on the synthetic instruction tuning dataset containing instruction-response pairs.
The model is trained using a cross entropy loss over the response tokens. 
Additional details in Section \ref{sec:experiments:setup}.

%% file: sections/experiments.tex
\section{Experiments}
\label{sec:experiments}

\subsection{Experiment Setup}\label{sec:experiments:setup}
\paragraph{Target Tasks and Datasets}
We consider three target tasks: yes-no question answering (YNQA), extractive question answering (ExQA), and natural language inference (NLI).
Table \ref{tab:exp:tasks} shows the seven datasets across three task types and the number of unannotated text in each dataset. 
We use the unannotated text from the datasets to train the specialized language models. 
For yes-no question answering, we choose PubMedQA~\citep{jin:emnlp19} and Privacy Policy QA~\citep{ravichander:emnlp19}. 
For extractive question answering, we choose the SquadShifts dataset~\citep{miller:icml20} that includes splits for the New York Times (NYT), Amazon, and Reddit. 
Finally, for the NLI task, we choose Contract-NLI~\citep{koreeda:emnlp21} and Vitamin C~\citep{schuster:acl21}.
We provide additional details in Appendix \ref{app:datasets}.

\input{tables/tasks}

In our experiments, we focus on tasks such as question answering and natural language inference that require us to generate a question or hypothesis and an answer. 
Prior work generates synthetic data for tasks like summarization that do not warrant a specialized task generation model~\citep{yehudai:iclr24}.
Other work focuses on generating instructions~\citep{li:arxiv23,koksal:arxiv23} for long-form text generation tasks where the solution to the instruction is the unannotated text. 
While these long-form generative tasks are useful for applications such as code generation, domains like biomedical and legal that we consider might benefit more from traditional predictive tasks~\citep{miller:cio2024}. 

\paragraph{Baselines}
We consider two key baselines: zero-shot and self supervised baseline.
For the zero-shot baseline, we prompt the model and run the evaluation without using any of the unannotated text from the target task (\textbf{None}). 
For the self supervised baseline, we use task-adaptive pretraining (\textbf{TAPT})~\citep{gururangan:acl20}. 
The learning objective is to continue to the pretraining objective on the unannotated text in the downstream dataset.
In our experiments, we use the next word prediction learning objective to fine-tune Mistral-7B and Llama 2 7B models. 

\paragraph{Synthetic Task Generation}
As described in Section \ref{sec:bonito:training}, we prompt \sys with the unannotated texts and the target task type to generate the instruction tuning data. 
We use nucleus sampling~\citep{holtzman:iclr2019} with a top P value of $0.95$ and a temperature of $0.5$, and a maximum sequence length of $256$ in the vLLM framework~\citep{kwon:sigops23}.

\paragraph{Models}
We adapt two pretrained large language models: Mistral-7B~\citep{jiang:arxiv23} and Llama 2 7B~\citep{touvron:arxiv23}.
They are decoder language models trained with the next word prediction objective on trillions of tokens. 
Both these models have around 7 billion parameters, with slightly different architectures optimized for sequence length and inference. 
For more details, see \citet{touvron:arxiv23} and \citet{jiang:arxiv23}.

We also consider a more practical setting where we further adapt instruction tuned models to the target task.
We first consider an off-the-shelf instruction tuned model: Mistral-7B-Instruct-v0.2. 
This model based on Mistral-7B achieves comparable performance to Llama 2 13B Chat on the MT-Bench~\citep{zheng:arxiv23}.
In addition, we train Mistral-7B and Llama 2 models on the T0 split from the P3 dataset~\citep{bach:acldemo22} and adapt them to the target tasks.  
We call these models Mistral-7B$_\mathrm{P3}$ and Llama 2$_\mathrm{P3}$.
For the instruction tuning details, see Appendix \ref{app:training_details:instruction_tuned}

\input{tables/base_results}
\input{tables/instruction_tuned_results}

\paragraph{Training Details}
We fine-tune the language models on the supervision sources---TAPT, and \sys---using Q-LoRA~\citep{dettmers:arxiv23}.
When further adapting Mistral-7B$_\mathrm{P3}$ and Llama 2 7B$_\mathrm{P3}$, we fine-tune the same Q-LoRA adapter on the supervision sources instead of merging and reinitializing the adapters. 
We train all the models for 1 epoch. 
If the dataset size is greater than 160,000 examples, then we train for 10,000 steps. 
We use the same hyperparameter values from \citet{dettmers:arxiv23} to avoid additional hyperparameter tuning.
Depending on the dataset, training on four GPUs takes 25 minutes to 17 hours. 
For more additional details, see Appendix
 \ref{app:training_details:hyperparameters}.

\paragraph{Evaluation}
We evaluate the performance of the models on the test splits of the target datasets (see Table \ref{tab:test_dataset} in Appendix \ref{app:datasets}).
To prevent ``prompt hacking'', following \citet{sanh:iclr22}, we first write five prompt templates for the target datasets and then benchmark the model performance. 
See Appendix \ref{app:prompt_eval} for all the prompts used in our experiments. 
We follow standard evaluation practices and report the F1 score for all the datasets. 
Following ~\citet{radford:gpt19}, we evaluate yes-no question answering and NLI using ranked classification, i.e., we generate the loglikelihood of all the choices and choose the sequence with the highest loglikelihood as the prediction. 
Following ~\citet{rajpurkar:emnlp2016}, we evaluate models on extractive question answering by computing the SQuAD F1 score on the generated output. 
During evaluation, we use greedy decoding to generate the output from the model and then calculate the SQuAD F1 score for the dataset. 

\subsection{Adapting Pretrained Models}\label{sec:experiments:base}
Table \ref{tab:experiment:base} shows that adapting pretrained models with synthetic instruction tuning data generated from \sys significantly outperforms zero-shot and TAPT.
\sys improves over the zero-shot performance by an average of 37.7 F1 points across Mistral-7B and Llama 2.
Although TAPT shows a nominal improvement of only 4.5 F1 points on average, we find that \sys outperforms TAPT by an average of 33.3 F1 points across both models. 
This result strengthens our main claim that synthetic instruction tuning data is a much better way of providing domain knowledge compared to self supervision. 
Finally, we observe that the Mistral-7B shows significantly greater improvement in performance compared to Llama 2 7B suggesting that stronger pretrained models might respond better to synthetic instructions. 

\input{tables/self_distillation}
\subsection{Adapting Instruction Tuned Models}\label{sec:experiments:instruction_tuned}
Table \ref{tab:experiment:instruct} shows that \sys improves instruction tuned models by an average of 22.1 F1 points whereas TAPT reduces the average performance by 0.8 F1 points.
This is because self supervision with TAPT interferes with prior instruction tuning and leads to catastrophic forgetting~\citep{french:cogsci99,kirkpatrick:pnas17}. 
In contrast, adapting instruction tuned models with \sys-generated tasks further improves performance in specialized domains. 
We also observe that \sys addresses the task-specific deficiencies and improves the instruction tuned models. 
For example, we find that \sys significantly improves Mistral-7B-Instruct-v0.2 performance on extractive question answering as it typically generates chat-like responses for questions. 
Finally, adapting instruction tuned variants of Mistral-7B and Llama 2 7B achieves a higher F1 score than adapting the pretrained models (Table \ref{tab:experiment:base}). 

%% file: tables/tasks.tex
\begin{table}[t]
    \centering
    \resizebox{1.0\linewidth}{!}{
    \begin{tabular}{llllr}\toprule
        \textbf{Task} && \textbf{Dataset} && \textbf{\# Unannotated} \\\cmidrule{1-1}\cmidrule{3-3}\cmidrule{5-5}
        \multirow{2}{*}{Yes-No QA} && PubmedQA && 211,269\\
        && Privacy Policy QA && 10,923\\\midrule
        \multirow{3}{*}{Extractive QA} && SquadShifts-NYT && 10,065\\
        && SquadShifts-Amazon && 9,885\\
        && SquadShifts-Reddit && 9,803\\\midrule
        \multirow{2}{*}{NLI} && Contract-NLI && 6,819\\
        && Vitamin C && 370,653\\\bottomrule
    \end{tabular}
    }
    \caption{Statistics of tasks and datasets used in the experiments.}
    \label{tab:exp:tasks}
\end{table}

%% file: tables/base_results.tex
\begin{table*}[t!]
    \centering
\resizebox{1.0\linewidth}{!}{
    \begin{tabular}{llllcclccclcclcc}\toprule
     && \multirow[b]{2}{*}{\parbox{1.5cm}{Supervision\\ Source}} && \multicolumn{2}{c}{Yes-No QA} && \multicolumn{3}{c}{Extractive QA} && \multicolumn{2}{c}{NLI} \\\cmidrule{5-6}\cmidrule{8-10}\cmidrule{12-13}
     Model &&  && PubMedQA & PrivacyQA && NYT & Amazon & Reddit && ContractNLI & Vitamin C && Average & $\Delta$\\\midrule
     \multirow{3}{*}{Mistral}&& None && 25.6 $_{2.1}$ & 44.1 $_{2.1}$ && 24.1 $_{1.6}$ & 17.5 $_{2.5}$ & 12.0 $_{2.6}$ && 31.2 $_{0.6}$ & 38.9 $_{0.6}$ && 27.6 & - \\
        && TAPT && 27.2 $_{2.3}$ & 46.3 $_{1.2}$ && 33.5 $_{4.3}$ & 25.5 $_{5.9}$ & 22.8 $_{7.0}$ && 34.2 $_{0.7}$ & 34.7 $_{2.6}$ && 32.0 & \increase{4.4} \\\cmidrule{3-16}
        && Bonito && \textbf{47.1} $_{1.0}$ & \textbf{52.5} $_{3.0}$ && \textbf{80.0} $_{1.0}$ & \textbf{72.5} $_{1.0}$ & \textbf{71.4} $_{1.6}$ && \textbf{71.9} $_{0.8}$ & \textbf{71.7} $_{0.2}$ && \textbf{66.7} & \increase{39.1} \\\midrule
     \multirow{3}{*}{Llama2}&& None && 23.7 $_{0.0}$ & 43.9 $_{3.0}$ && 20.1 $_{2.4}$ & 14.4 $_{2.0}$ & 11.0 $_{1.9}$ && 28.6 $_{2.2}$ & 22.2 $_{2.9}$ && 23.4 & -\\
        && TAPT &&  23.7 $_{0.0}$ & 44.1 $_{2.3}$ && 26.7 $_{6.6}$ & 25.4 $_{5.9}$ & 20.6 $_{6.8}$ && 29.8 $_{2.4}$ & 26.2 $_{2.0}$ && 28.1 & \increase{4.6}\\\cmidrule{3-16}
        && Bonito && \textbf{26.1} $_{2.1}$ & \textbf{51.4} $_{2.2}$ && \textbf{75.3} $_{1.9}$ & \textbf{66.5} $_{1.9}$ & \textbf{63.7} $_{3.0}$ && \textbf{63.9} $_{1.1}$ & \textbf{70.7} $_{0.5}$ && \textbf{59.7} & \increase{36.2}\\\bottomrule
    \end{tabular}
    }
    \caption{
    Results for zero-shot task adaptation with pretrained base models. 
    We report the F1 and the standard error averaged across five prompt templates for all the datasets.
    }
    \label{tab:experiment:base}
\end{table*}

%% file: tables/instruction_tuned_results.tex
\begin{table*}[t!]
    \centering
\resizebox{1.0\linewidth}{!}{
    \begin{tabular}{llllcclccclcclcc}\toprule
     && \multirow[b]{2}{*}{\parbox{1.5cm}{Supervision\\ Source}} && \multicolumn{2}{c}{Yes-No QA} && \multicolumn{3}{c}{Extractive QA} && \multicolumn{2}{c}{NLI} \\\cmidrule{5-6}\cmidrule{8-10}\cmidrule{12-13}
     Model &&  && PubMedQA & PrivacyQA && NYT & Amazon & Reddit && ContractNLI & Vitamin C && Average & $\Delta$\\\midrule
    \multirow{3}{*}{
    \parbox{2cm}{Mistral-7B-Instruct-v0.2}}&& None && 32.8 $_{0.3}$ & \textbf{57.9} $_{2.9}$ && 19.7 $_{2.7}$ & 15.8 $_{2.4}$ & 13.0 $_{2.2}$ && 55.4 $_{2.0}$ & 58.0 $_{1.1}$ && 36.1 & - \\
        && TAPT && 28.3 $_{0.5}$ & 56.3 $_{2.4}$ && 37.9 $_{2.2}$ & 30.1 $_{2.2}$ & 26.3 $_{4.6}$ && 42.5 $_{1.8}$ & 49.6 $_{1.8}$ && 38.7 & \increase{2.6}\\\cmidrule{3-16}
        && Bonito && \textbf{41.7} $_{0.4}$ & 56.2 $_{3.5}$ && \textbf{80.1} $_{1.0}$ & \textbf{72.8} $_{1.1}$ & \textbf{71.8} $_{1.4}$ && \textbf{70.9} $_{1.8}$ & \textbf{72.6} $_{0.1}$ && \textbf{66.6} & \increase{30.5} \\\midrule
     \multirow{3}{*}{Mistral-7B$_{\mathrm{P3}}$}&& None && 45.1 $_{1.3}$ & 49.9 $_{2.6}$ && 73.8 $_{0.8}$ & 61.0 $_{2.3}$ & 60.6 $_{2.2}$ && 33.3 $_{0.7}$ & 46.0 $_{0.6}$ && 52.8 & - \\
        && TAPT && \textbf{51.1} $_{2.2}$ & 42.8 $_{3.7}$ && 70.8 $_{1.7}$ & 59.7 $_{3.2}$ & 58.0 $_{2.6}$ && 38.1 $_{3.6}$ & 43.6 $_{0.4}$ && 52.0 & \decrease{0.8}\\\cmidrule{3-16}
        && Bonito && 46.1 $_{0.5}$ & \textbf{56.7} $_{4.3}$ && \textbf{80.7} $_{0.7}$ & \textbf{73.9} $_{0.6}$ & \textbf{72.3} $_{1.1}$ && \textbf{71.8} $_{0.5}$ & \textbf{73.9} $_{0.1}$ && \textbf{67.9} & \increase{15.1} \\\midrule
     \multirow{3}{*}{Llama 2$_{\mathrm{P3}}$}&& None && 26.0 $_{0.5}$ & 38.5 $_{1.9}$ && 64.2 $_{2.6}$ & 50.6 $_{3.6}$ & 49.4 $_{4.1}$ && 23.5 $_{2.6}$ & 44.6 $_{0.3}$ && 42.4 & -\\
        && TAPT && 25.1 $_{0.6}$ & 42.0 $_{3.8}$ && 51.4 $_{6.7}$ & 47.0 $_{4.8}$ & 42.2 $_{5.8}$ && 22.6 $_{3.0}$ & 36.9 $_{1.7}$ && 38.2 & \decrease{4.4}\\\cmidrule{3-16}
        && Bonito && 
        \textbf{27.0} $_{1.7}$ & \textbf{56.9} $_{3.8}$ && \textbf{77.5} $_{1.4}$ & \textbf{69.6} $_{1.1}$ & \textbf{68.2} $_{1.9}$ && \textbf{68.5} $_{0.7}$ & \textbf{73.7} $_{0.3}$ && \textbf{63.1} & \increase{20.7}\\\bottomrule
    \end{tabular}
    }
    \caption{
    Results for zero-shot task adaptation of instruction tuned models. 
    We report the F1 and the standard error averaged across five prompt templates for all the datasets.
    }
    \label{tab:experiment:instruct}
\end{table*}

%% file: tables/self_distillation.tex
\begin{table*}[t!]
    \centering
\resizebox{1.0\linewidth}{!}{
    \begin{tabular}{llllcclccclcclcc}\toprule
     && \multirow[b]{2}{*}{\parbox{1.5cm}{Supervision\\ Source}} && \multicolumn{2}{c}{Yes-No QA} && \multicolumn{3}{c}{Extractive QA} && \multicolumn{2}{c}{NLI} \\\cmidrule{5-6}\cmidrule{8-10}\cmidrule{12-13}
     Model &&  && PubMedQA & PrivacyQA && NYT & Amazon & Reddit && ContractNLI & Vitamin C && Average & $\Delta$\\\midrule
     \multirow{3}{*}{\parbox{3.25cm}{Mistral-7B-Instruct-v0.2$_{\mathrm{special}}$}} && None && 47.5 $_{0.3}$ & 59.1 $_{1.5}$ && \textbf{82.6} $_{0.5}$ & \textbf{77.6} $_{0.7}$ & \textbf{75.6} $_{0.8}$ && \textbf{77.3} $_{0.1}$ & 70.3 $_{0.1}$ && 70.0 & -\\
     && Bonito && 47.4 $_{0.2}$ & \textbf{62.3} $_{0.9}$ && 82.4 $_{0.6}$ & 76.0 $_{0.6}$ & 74.9 $_{0.9}$ && 75.1 $_{1.0}$ & 71.9 $_{0.1}$ && 70.0 & \increase{0.0}\\
     && Bonito$_\mathrm{special}$ && \textbf{50.3} $_{0.1}$ & 59.8 $_{1.3}$ && 81.8 $_{0.7}$ & 76.4 $_{0.8}$ & 74.5 $_{1.0}$ && 77.0 $_{0.4}$ & \textbf{73.5} && \textbf{70.5} & \increase{0.5} \\\midrule
     \multirow{3}{*}{Mistral-7B$_{\mathrm{special}}$} && None && 36.7 $_{1.9}$ & 54.4 $_{1.4}$ && \textbf{82.6} $_{0.5}$ & \textbf{76.6} $_{0.8}$ & 75.0 $_{0.8}$ && 75.1 $_{0.3}$ & 71.8 $_{0.2}$ && 67.5 & -\\
     && Bonito && 42.7 $_{1.2}$ & 55.1 $_{1.7}$ && 82.5 $_{0.4}$ & 76.1 $_{0.6}$ & 74.3 $_{1.1}$ && 76.7 $_{0.2}$ & 71.4 $_{0.1}$ && 68.4 & \increase{0.9}\\
     && Bonito$_\mathrm{special}$ && \textbf{49.3} $_{0.4}$ & \textbf{57.2} $_{1.6}$ && 81.7 $_{0.8}$ & 76.2 $_{0.8}$ & \textbf{75.3} $_{0.9}$ && \textbf{76.8} $_{0.2}$ & \textbf{73.8} $_{0.1}$ && \textbf{70.0} & \increase{2.5}\\\midrule
    \end{tabular}
    }
    \caption{
    Results for adapting task-specialized models on the downstream target datasets. 
    We report the F1 and the standard error averaged across five prompt templates for all the datasets. 
    }
    \label{tab:experiment:self}
\end{table*}

%% file: sections/analysis.tex
\section{Analysis}
\subsection{Impact of Domain Knowledge}\label{sec:analysis:self_distillation}
Here we ask a key question: are we improving the language model by learning about the domain or are we distilling instructing tuning data from a stronger to a weaker model?
To answer this question, we train task-specialized instruction tuned models and then further train them on synthetic tasks generated from \sys for the target unannotated texts. 
We create the task-specialized training dataset by selecting the instructions in CTGA with the target task type.
We train two task-specialized models in the standard instruction-response format: Mistral-7B-Instruct-v0.2$_\mathrm{special}$ and Mistral-7B$_\mathrm{special}$. 
We also train a task-specialized \sys$_{\mathrm{special}}$ on the same task-specific dataset. 
See Appendix \ref{app:training_details:spl_bonito} for training details. 

Table \ref{tab:experiment:self} 
shows that further training on synthetic instructions can improve performance suggesting that the model benefits from the unnannotated text from the specialized domain. 
We find that training on \sys tasks either slightly improves or matches the performance of task-specialized models on average. 
When we train on \sys$_{\mathrm{special}}$ tasks, we further improve task-specialized Mistral-7B-Instruct-v0.2 by 0.5 F1 points and Mistral-7B and 2.5 F1 points. 
We see that the model performance often reduces on extractive QA.
We suspect that the model performance has saturated due to the presence of SQuAD in the task-specialized training dataset. 
To further improve on extractive question answering, we could benefit from having access to a few examples from the target dataset. 
Finally, we almost always improve performance on Vitamin C and PubMedQA datasets highlighting the importance of training on more task samples (see Section \ref{sec:analysis:training_size}). 
\input{fig/training_size}

\vspace{-1.em}
\subsection{Effect of the Training Dataset Size}\label{sec:analysis:training_size}
Here we study the effect of the size of the training dataset.
In particular, we study how Mistral-7B performance varies when trained on different quantities of synthetic instruction tuning data for PubMedQA and Vitamin C.
Figure \ref{fig:size_of_training} shows that training on more steps typically improves performance. 
We find that \sys on PubMedQA reaches the peak performance of 47.1 F1 points after 10,000 steps but the F1 can fluctuate when trained for fewer steps. 
In contrast, we find that \sys gets the highest performance of 73.3 F1 points after 2,500 points and gradually diminishes the performance to 71.7 F1 points. 
Finally, if available, we suggest using a validation set to select the best-performing model checkpoint.

\subsection{Human Evaluation: Agreement with GPT-4o}
We manually evaluate Bonito tasks by comparing the answers generated by Bonito and GPT-4o. 

\input{tables/human_eval}
\paragraph{Setup}
We sample 100 unique instructions each from Bonito for PubMedQA, SQuADShifts Reddit, and ContractNLI.
Next, we prompt GPT-4o with instructions generated by Bonito to produce an answer. 
We prefix the instructions with a simple format prompt to produce answers in the desired format with GPT-4o.
Finally, we ask humans if GPT-4o and Bonito produce the same (including paraphrased) answers for the instructions.
For reproducibility, we use GPT-4o-2024-05-13.
We separately ask the first three authors of the paper to compare the responses from both models. 
We choose the final agreement if two or more annotators agree on either a match or no match.

\paragraph{Results}
Table \ref{tab:human_eval} shows that Bonito and GPT-4o produce the same answer for Bonito tasks 71  to 76 percent of the time across three datasets, with high inter-annotator agreement. 
Each dataset reveals different patterns of disagreement.  
In PubMedQA, GPT-4o generates the response as ``unanswerable'' when the answer is not present in the passage, whereas Bonito produces either ``yes'' or ``no'', or ``true'' or ``false''. 
In SQuADShifts Reddit, Bonito almost always extracts the answer from the paragraph, whereas GPT-4o can generate answers with additional text that may not be present in the paragraph. 
In ContractNLI, both models can produce plausible answers. 
In one example, Bonito generates the question, `Does this imply that ``This document is confidential information''? Yes, no, or maybe?'.
Bonito answers ``yes'', whereas GPT-4o produces the answer ``maybe''. 
In such cases, we annotate the responses as no match, reducing the agreement. 

Our analysis shows that Bonito generates high-quality tasks with accurate answers. 
However, there is still room to improve the quality of the tasks. 
Improving the task quality could further increase the downstream model performance. 
Therefore, we believe that research on conditional task generation is an important direction for future work.

%% file: fig/training_size.tex
\begin{figure}[t]
    \centering
    \includegraphics[width=1.0\linewidth]{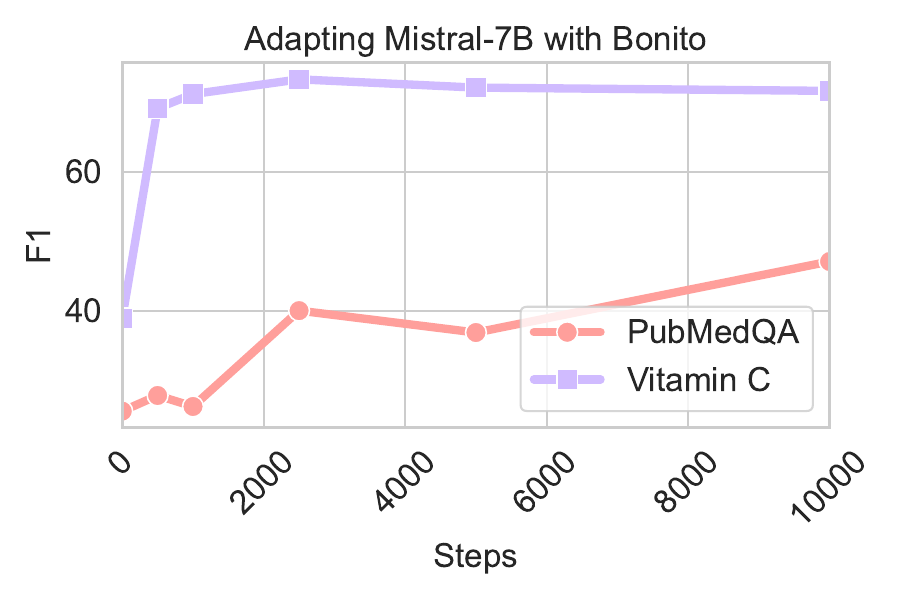}
    
    \caption{
    Adapting Mistral-7B with Bonito-generated tasks and evaluating performance after training for different number of steps.
    }
    \label{fig:size_of_training}
\end{figure}

%% file: tables/human_eval.tex
\begin{table}[t]
    \centering
    \resizebox{1.0\linewidth}{!}{
    \begin{tabular}{llclc}\toprule
        \textbf{Dataset} && \textbf{Match=Yes} && \textbf{Agreement (>=2)}\\\cmidrule{1-1}\cmidrule{3-3}\cmidrule{5-5}
       PubMedQA  && 72\% && 97\% \\
       Reddit  && 76\% && 88\%\\
       ContractNLI && 71\% && 99\%\\\bottomrule
    \end{tabular}
    }
    \caption{Agreement between GPT-4o and Bonito generated answers for Bonito tasks. 
    Agreement (>=2) is the agreement percentage when two or more annotators agree on a match or no match.}
    \label{tab:human_eval}
\end{table}

%% file: sections/additional.tex
\section{Additional Experiments}\label{sec:additional}
We briefly describe additional experiments included in Appendix \ref{app:openly} and \ref{app:gpt4}. 

In Appendix \ref{app:openly}, we generate synthetic tasks by prompting Mistral-7B-Instruct-v0.2 and Zephyr-7B-$\beta$.
Our results show that the synthetic tasks from Mistral-7B-Instruct-v0.2 and Zephyr-7B-$\beta$ improve the average performance of Mistral-7B but decrease significantly when adapting Mistral$_{\mathrm{P3}}$. 
This indicates that we require high-quality synthetic tasks to increase the performance of strong instruction-tuned models.
In Appendix \ref{app:gpt4}, we generate synthetic tasks with GPT-4 for Privacy Policy QA, SQuADShifts Reddit, and ContractNLI.
Our results show that GPT-4 improves Mistral$_{P3}$ on Privacy Policy QA and ContractNLI but slightly reduces performance on SQuADShifts Reddit. 
Finally, we analyze the generated tasks and identify common issues with both open-source models and GPT-4, such as the distribution of the label space and ``chatty'' responses, which potentially lead to the drop in performance.

%% file: sections/conclusion.tex
\section{Conclusion}
We present Bonito, an open-source model for conditional task generation that converts unannotated text into instruction tuning datasets.
We show that training with synthetic instruction tuning datasets in specialized domains is a strong alternative to self supervision. 
Our experiments demonstrate that \sys-generated instructions improve pretrained and instruction tuned models on zero-shot task adaptation. 
Overall, \sys enables practitioners to adapt large language models to tasks on their data without annotations.

%% file: appendix/appendix.tex
\input{appendix/datasets}

\input{appendix/openly_available_models}
\input{appendix/gpt4}
\input{appendix/flan}
\input{appendix/pythia}
\input{appendix/training_details}

\input{appendix/use_of_ai}
\input{appendix/ctga}

\input{appendix/target_prompts}

\input{appendix/qualitative}

%% file: appendix/datasets.tex
\section{Datasets}\label{app:datasets}
We briefly describe the datasets used in our experiments.
We get all the datasets from the datasets library \citep{lhoest:emnlp21}.
For all the datasets, we consider five prompt templates (see Appendix \ref{app:prompt_eval}).
Table \ref{tab:test_dataset} shows the statistics for the test splits in the evaluation datasets. 
Below we include details about the evaluation datasets: 
\begin{itemize}
    \item \textbf{PubMedQA}~\citep{jin:emnlp19}: The dataset asks questions about PubMed abstracts that can be answered with a yes, no, or maybe. 
    We use the abstracts without the questions as unannotated text for adaptation. 
    During the evaluation, we provide the PubMed abstract along with the question from the test set to the model. 
    \item \textbf{Privacy Policy QA}~\citep{ravichander:emnlp19}: The dataset consists of paragraphs from privacy policies with corresponding questions. 
    The task involves determining the relevance of each question, formatted as a yes-no question-answering task.
    We use the processed test split in Privacy Policy QA from ~\citet{guha:arxiv23} as unannotated text.
    
    \item \textbf{SquadShifts}~\citep{miller:icml20}: The dataset is designed to test the robustness of extractive question answering models. 
    We use three of the four test sets in our work --- New York Times articles, Reddit posts, and Amazon product reviews. 
    During training, we use the articles or context from the test set without the questions and generate extractive question answering tasks with \sys. 
    During evaluation, we evaluate the same test set with the questions in the dataset. 
    \item \textbf{ContractNLI}~\citep{koreeda:emnlp21}: ContractNLI is a natural language inference task to aid contract review. 
    Given a hypothesis about a clause in a contract, the model predicts if the hypothesis is supported, refuted, or not mentioned.
    \item \textbf{Vitamin C}~\citep{schuster:acl21}: This dataset focuses on fact verification in Wikipedia framed as a natural language inference task. 
    Each example consists of an evidence text from Wikipedia and a corresponding fact. 
    The model is asked to indicate whether the fact is supported, refuted, or neutral. 
    
\end{itemize}

\input{tables/test_dataset}

%% file: tables/test_dataset.tex
\begin{table}[t]
    \centering
    \resizebox{1.0\linewidth}{!}{
    \begin{tabular}{llrlr}\toprule
        \textbf{Dataset} && \textbf{\# Classes} && \textbf{\# Test Examples} \\\cmidrule{1-1}\cmidrule{3-3}\cmidrule{5-5}
        PubmedQA && 3 && 500\\
        Privacy Policy QA && 2 && 10,923\\\midrule
        SquadShifts-NYT && - && 10,065\\
        SquadShifts-Amazon && - && 9,885\\
        SquadShifts-Reddit && - && 9,803\\\midrule
        Contract-NLI && 3 && 1,991\\
        Vitamin C && 3 && 55,197\\\bottomrule
    \end{tabular}
    }
    \caption{Statistics for the evaluation test sets in the datasets from our experiments.
    ``-'' in the number of classes indicates a generation task. 
    }
    \label{tab:test_dataset}
\end{table}

%% file: appendix/openly_available_models.tex
\input{tables/prompts}
\input{tables/openly_available_results}

\section{Generating Tasks with Open-Source Models}\label{app:openly}
We use Mistral-Instruct-v0.2 and Zephyr-$\beta$, two popular openly available models, to generate instruction tuning data. 
Then, we adapt pretrained Mistral-7B and Mistral-7B-$_\mathrm{P3}$ on the generated tasks. 

\subsection{Generating Synthetic Datasets}\label{app:openly:gen_synth}
Here we describe the process of creating synthetic datasets with Mistral-Instruct-v0.2 and Zephyr-$\beta$.
We prompt these models to generate questions or hypotheses for the target unannotated text.
Table \ref{tab:prompts} shows the prompts we used to generate the tasks. 
Creating these prompts required a tremendous amount of prompt engineering as they struggled to follow the prompt format~\citep{xia:arxiv24}. 
We first generate the question or the hypothesis and then re-prompt the model to produce the answer. 
For question answering tasks, we prepend the question as the prompt followed by the unannotated text to generate the output. 
For the NLI datasets, we use five prompt templates from the ANLI dataset in \citet{bach:acldemo22} and plug in the hypothesis and the unannotated text as the input to the model to generate the answer.
We use the same input and output to adapt the pretrained and instruction tuned models. 
For all the generations, we use a top-P of 0.95, temperature of 0.5, and maximum token length of 256. 

\subsection{Results}
Table \ref{tab:experiment:openly_available} shows results for zero-shot task adaptation with openly available models. 
We see that both Mistral-7B-Instruct-v0.2 and Zephyr-7B-$\beta$ improve performance over the pretrained Mistral-7B but find that they severely hurt average performance compared to Mistral-7B$_\mathrm{P3}$.

We suspect that the drop in performance is due to issues related to the generated tasks.
For extractive question answering, we find that Mistral-7b-Instruct-v0.2 and Zephyr-$\beta$ often generate questions with multiple sub-questions that cannot be easily answered by extracting words from the context.
Furthermore, the responses are ``chatty'' which might not be appropriate for extractive question answering.
We also observe that the generated questions are often ``positive'', i.e., they usually have ``yes'' or ``true'' as the answer.
For example, 68\% of the questions generated by Zephyr-$\beta$ for PubMedQA have answers starting with ``yes'' or ``true'' but only 5\% of the questions have answers that start with ``no'' or ``false''. 
We observe a similar ``positive'' bias in the hypotheses generated for natural language inference datasets.

%% file: tables/prompts.tex
\begin{table}[]
    \centering
    \begin{tabular}{c}\toprule
    \begin{minipage}{\linewidth}
    \begin{small}
    \begin{lmttfont}
    \underline{Task Type:} Yes-no question answering\\\\
    \underline{Prompt:} Generate exactly one question that can be answered by a yes or a no for the paragraph below. The question should be parsable and enclosed in quotes (""). \\
    <context>
    \end{lmttfont}
    \end{small}
    \end{minipage}\\\midrule
    \begin{minipage}{\linewidth}
    \begin{small}
    \begin{lmttfont}
    \underline{Task Type:} Extractive question answering\\\\
    \underline{Prompt:} Generate exactly one question that can be answered by selecting 1 to 10 words from the paragraph below. The question should be parsable and enclosed in quotes ("").\\
    <context>
    \end{lmttfont}
    \end{small}
    \end{minipage}\\\midrule
    \begin{minipage}{\linewidth}
    \begin{small}
    \begin{lmttfont}
    \underline{Task Type:} Natural language inference\\\\
    \underline{Prompt:} Generate exactly one high-level statement or a hypothesis for the following paragraph. The hypothesis about the paragraph can be true, false, or neither. Make sure the output is less than 10 words. The hypothesis should be parsable and enclosed in quotes (""). \\
    <context>
    \end{lmttfont}
    \end{small}
    \end{minipage}\\\midrule
    \end{tabular}
    \caption{Prompts used generated tasks with Mistral-Instruct-v0.2, Zephyr-$\beta$, and GPT-4.
    We replace \texttt{<context>} with the unannotated text. }
    \label{tab:prompts}
\end{table}

%% file: tables/openly_available_results.tex
\begin{table*}[t!]
    \centering
\resizebox{1.0\linewidth}{!}{
    \begin{tabular}{llllcclccclcclcc}\toprule
     && \multirow[b]{2}{*}{\parbox{1.5cm}{Supervision\\ Source}} && \multicolumn{2}{c}{Yes-No QA} && \multicolumn{3}{c}{Extractive QA} && \multicolumn{2}{c}{NLI} \\\cmidrule{5-6}\cmidrule{8-10}\cmidrule{12-13}
     Model &&  && PubMedQA & PrivacyQA && NYT & Amazon & Reddit && ContractNLI & Vitamin C && Average & $\Delta$\\\midrule
     \multirow{4}{*}{Mistral-7B} && None && 25.6 $_{2.1}$ & 44.1 $_{2.1}$ && 24.1 $_{1.6}$ & 17.5 $_{2.5}$ & 12.0 $_{2.6}$ && 31.2 $_{0.6}$ & 38.9 $_{0.6}$ && 27.6 & -\\
     && Mistral-Instruct-v0.2 && 29.4 $_{0.8}$ & 50.1 $_{5.5}$ && 22.3 $_{1.7}$ & 17.2 $_{1.9}$ & 13.6 $_{2.1}$ && 55.3 $_{1.4}$ & 52.2 $_{1.5}$ && 34.3 & \increase{6.7}\\
     && Zephyr-$\beta$ && 32.2 $_{1.6}$ & \textbf{59.4} $_{2.3}$ && 20.4 $_{1.5}$ & 18.2 $_{1.9}$ & 15.0 $_{2.1}$ && 33.3 $_{2.9}$ & 51.9 $_{3.0}$ && 32.9 & \increase{5.3} \\\cmidrule{3-16}
     && Bonito && \textbf{47.1} $_{1.0}$ & 52.5 $_{3.0}$ && \textbf{80.0} $_{1.0}$ & \textbf{72.5} $_{1.0}$ & \textbf{71.4} $_{1.6}$ && \textbf{71.9} $_{0.8}$ & \textbf{71.7} $_{0.2}$ && \textbf{66.7} & \increase{39.1}\\\bottomrule
     \multirow{4}{*}{Mistral-7B$_{\mathrm{P3}}$} && None && 45.1 $_{1.3}$ & 49.9 $_{2.6}$ && 73.8 $_{0.8}$ & 61.0 $_{2.3}$ & 60.6 $_{2.2}$ && 33.3 $_{0.7}$ & 46.0 $_{0.6}$ && 52.8 & -\\
     && Mistral-Instruct-v0.2 && 34.1 $_{1.1}$ & \textbf{62.1} $_{1.4}$ && 24.1 $_{1.7}$ & 18.8 $_{2.2}$ & 15.3 $_{2.2}$ && 53.9 $_{1.8}$ & 53.5 $_{1.0}$ && 37.4 & \decrease{15.4}\\
     && Zephyr-$\beta$ && 38.8 $_{1.7}$ & 55.3 $_{3.5}$ && 22.2 $_{1.6}$ & 20.0 $_{2.0}$ & 16.6 $_{2.0}$ && 36.5 $_{5.7}$ & 51.6 $_{3.2}$ && 34.4 & \decrease{18.4}\\\cmidrule{3-16}
     && Bonito && \textbf{46.1} $_{0.5}$ & 56.7 $_{4.3}$ && \textbf{80.7} $_{0.7}$ & \textbf{73.9} $_{0.6}$ & \textbf{72.3} $_{1.1}$ && \textbf{71.8} $_{0.5}$ & \textbf{73.9} $_{0.1}$ && \textbf{67.9} & \increase{15.1}\\\bottomrule
    \end{tabular}
    }
    \caption{
    Results for zero-shot task adaptation with tasks generated from Mistral-Instruct-v0.2 and Zephyr-$\beta$.
    We report the F1 and the standard error averaged across five prompt templates for all the datasets. 
    }
    \label{tab:experiment:openly_available}
\end{table*}

%% file: appendix/gpt4.tex
\input{tables/gpt4_results}
\input{tables/flan}

\section{Generating Tasks with GPT-4}\label{app:gpt4}
Here we use GPT-4 to generate tasks to adapt Mistral-7B$_\mathrm{P3}$. 

\subsection{Generating Synthetic Datasets}\label{app:gpt_4:gen_synth}
We prompt GPT-4 to generate tasks for Privacy Policy QA, SQuADShifts Reddit, and Contract NLI. 
For simplicity, we use the same prompts from Appendix \ref{app:openly:gen_synth} to generate questions and hypotheses (see Table \ref{tab:prompts}). 
For Privacy Policy QA, we add a simple instruction prefix to answer the question with yes or no along with the question and the context to generate the answer.
For extractive question answering, we add the prefix "Extract the exact words from the paragraph for the question. 
If the question is not answerable, say N/A." before the question and the context and produce the answer.
We use a simpler prefix "Answer the following question." when training the downstream model on SQuADShifts Reddit.
Finally, for ContractNLI, we use the same prompts from Appendix \ref{app:openly:gen_synth} to generate answers. 
For all the generations, we use gpt-4-0613 with a maximum token length of 256, top-P of 0.95, and temperature of 0.5.

\subsection{Results}
Table \ref{tab:results:gpt} shows that tasks generated by GPT-4 improve performance over Mistral-7B$\mathrm{P3}$ on Privacy Policy QA and ContractNLI but slightly reduce performance on SQuADShifts Reddit. 
While GPT-4 is a much better task generator than the open-source models, we find that GPT-4 also suffers from a similar issue. 
For example, ContractNLI often has a positive hypothesis and PrivacyQA has a question with the answer yes.
While GPT-4 follows the instruction to generate exactly one question for the paragraph, we find that it produces slightly longer answers to the question. 
The SQuAD metric penalizes if there unwanted tokens in the answers. 
Finally, the cost of generating tasks with GPT-4 makes it prohibitively expensive to generate tasks for larger datasets like PubMedQA and Vitamin C. 

%% file: tables/gpt4_results.tex
\begin{table}[t!]
    \centering
\resizebox{1.0\linewidth}{!}{
    \begin{tabular}{llllcclccclcclcc}\toprule
    Model & Sup. src.  & PrivacyQA & Reddit & ContractNLI \\\midrule
    \multirow{3}{*}{Mistral-7B$_{\mathrm{P3}}$} & None & 49.9 $_{2.6}$ & 61.0 $_{2.8}$ & 33.3 $_{0.7}$\\
    & GPT-4 & \textbf{57.2} $_{4.8}$ & 52.4 $_{3.0}$ & 43.1 $_{0.7}$ \\\cmidrule{2-5}
    & Bonito & 56.7 $_{4.3}$ & \textbf{72.3} $_{1.1}$ & \textbf{71.8} $_{0.5}$\\\bottomrule
    \end{tabular}
    }
    \caption{
    Results for zero-shot task adaptation with task generated from GPT-4. 
    We report the F1 and the standard error averaged across five prompts templates for all the datasets.
    }
    \label{tab:results:gpt}
\end{table}

%% file: tables/flan.tex
\begin{table*}[t!]
    \centering
\resizebox{1.0\linewidth}{!}{
    \begin{tabular}{llcclccclcclc}\toprule
      && \multicolumn{2}{c}{Yes-No QA} && \multicolumn{3}{c}{Extractive QA} && \multicolumn{2}{c}{NLI} \\\cmidrule{3-4}\cmidrule{6-8}\cmidrule{10-11}
     Model && PubMedQA & PrivacyQA && NYT & Amazon & Reddit && ContractNLI & Vitamin C && Average\\\midrule
        FLAN-T5-XXL (11B) && 50.0 $_{0.4}$ & \textbf{62.5} $_{2.2}$ && \textbf{84.2} $_{0.2}$ & 72.3 $_{1.9}$ & 70.1 $_{3.1}$ && 45.4 $_{3.5}$ & 62.5 $_{2.7}$ && 63.9 \\
        FLAN-T5-XL (3B) && \textbf{52.5} $_{0.2}$ & 59.3 $_{1.6}$ && 82.1 $_{1.3}$ & 68.1 $_{5.4}$ & 67.3 $_{3.1}$ && 37.0 $_{0.6}$ & 54.7 $_{0.4}$ && 60.2 \\
        Mistral-7B-Instruct-v0.2 + Bonito && 41.7 $_{0.4}$ & 56.2 $_{3.5}$ && 80.1 $_{1.0}$ & 72.8 $_{1.1}$ & 71.8 $_{1.4}$ && 70.9 $_{1.8}$ & 72.6 $_{0.1}$ && 66.6 \\
        Mistral-7B$_{\mathrm{P3}}$ + Bonito && 46.1 $_{0.5}$ & 56.7 $_{4.3}$ && 80.7 $_{0.7}$ & \textbf{73.9} $_{0.6}$ & \textbf{72.3} $_{1.1}$ && \textbf{71.8} $_{0.5}$ & \textbf{73.9} $_{0.1}$ && \textbf{67.9} \\\bottomrule
    \end{tabular}
    }
    \caption{
    Results comparing zero-shot task adaptation of instruction tuned models with FLAN-T5 models. 
    We report the F1 and the standard error averaged across five prompt templates for all the datasets.
    }
    \label{tab:flan}
\end{table*}

%% file: appendix/flan.tex
\section{Bonito vs. FLAN}\label{app:flan}
We evaluate the zero-shot performance of FLAN-T5-XXL (11B) and FLAN-T5-XL (3B) models~\citep{longpre:icml23} on the target datasets used in our experiments. 
Table \ref{tab:flan} shows that Mistral-7B-Instruct-v0.2 and Mistral$_{\mathrm{P3}}$ with \sys-generated tasks improves over FLAN-T5-XXL (11B) by 2.7 F1 points and 4.0 F1 points. 
Our results also show that Mistral-7B-Instruct-v0.2 and Mistral$_{\mathrm{P3}}$ with \sys outperforms FLAN-T5-XL (3B) by 6.4 F1 points and 7.7 F1 points. 

%% file: appendix/pythia.tex
\section{Bonito with Smaller Models}
We report an additional comparison with Bonito trained on Pythia (2.8B) ~\citep{biderman:icml23}. 
We follow the same experimental setup used in Section \ref{sec:experiments:setup}.

\input{tables/pythia_results}

\paragraph{Results}
Table \ref{tab:pythia} shows that Bonito improves Pythia (2.8B) by an average of 30.3 F1 points across all the datasets. 
We observe that Pythia (2.8B) with Bonito performs better than Mistral with TAPT and Llama 2 with TAPT despite being twice as small (See Table \ref{tab:experiment:base}).
These results show that Bonito can be used to  
create small but powerful specialized language models.  

%% file: tables/pythia_results.tex
\begin{table*}[t!]
    \centering
\resizebox{1.0\linewidth}{!}{
    \begin{tabular}{llcclccclcclc}\toprule
      && \multicolumn{2}{c}{Yes-No QA} && \multicolumn{3}{c}{Extractive QA} && \multicolumn{2}{c}{NLI} \\\cmidrule{3-4}\cmidrule{6-8}\cmidrule{10-11}
     Model && PubMedQA & PrivacyQA && NYT & Amazon & Reddit && ContractNLI & Vitamin C && Average\\\midrule
    Pythia (2.8B) && 23.7 $_{0.0}$ & 42.2 $_{1.4}$ && 11.9 $_{0.9}$ & 8.9 $_{0.5}$ & 8.0 $_{0.6}$ && 20.8 $_{3.5}$ & 25.4 $_{1.5}$ && 20.1\\
    Pythia (2.8B) + Bonito && \textbf{25.9} $_{2.2}$ & \textbf{51.6} $_{0.9}$ && \textbf{59.8} $_{4.2}$ & \textbf{52.2} $_{3.5}$ & \textbf{51.7} $_{4.3}$ && \textbf{48.4} $_{2.5}$ & \textbf{63.3} $_{0.9}$ && \textbf{50.4}\\\bottomrule
    \end{tabular}
    }
    \caption{
    Results for pretrained Pythia and Pythia adapted with Bonito. 
    We report the F1 and the standard error averaged across five prompt templates for all the datasets.
    }
    \label{tab:pythia}
\end{table*}

%% file: appendix/training_details.tex
\section{Training Details}\label{app:training_details}
Here we provide training details for models used in the paper.

\subsection{Training \sys}\label{app:hyp:bonito}
We train Mistral-7B on the conditional task generation with attributes (CTGA) dataset.
From the training set, we uniformly sample 10,000 examples as the validation set to monitor the loss.
The rest of the dataset is used for training \sys. 
We train the model using Q-LoRA~\citep{dettmers:arxiv23} by optimizing the cross entropy loss over the output tokens. 
The model is trained for 100,000 steps. 
The training takes about 4 days on four GPUs to complete. 
We include all the hyperparameters in Appendix \ref{app:training_details:hyperparameters}.

The same training recipe can be used to train other existing language models such as Falcon~\citep{falcon40b}, Pythia ~\citep{biderman:icml23}, and RedPajama~\citep{together:github23}.
While models such as Llama 2 \citep{touvron:arxiv23} can be trained on CTGA, their license prohibits the use of the output to enhance any other large language model.

\subsection{Instruction Tuned Models}\label{app:training_details:instruction_tuned}
Here we describe the procedure to train Mistral-7B$_\mathrm{P3}$ and Llama 2 7B$_\mathrm{P3}$. 
We use the processed T0 dataset from ~\citet{muennighoff:acl23}.
Since the dataset is large, we uniformly sample 1.6 million input-output examples and train the language model on them.
Following \citet{dettmers:arxiv23}, we train the model for 10,000 steps with Q-LoRA and optimize the cross entropy loss over the output tokens. 
The training takes about 10 hours on four GPUs to complete. 
For the rest of the hyperparameters, see Appendix \ref{app:training_details:hyperparameters}.

\input{tables/hyperparameters}
\input{tables/task_distribution}
\subsection{Training Task-Specialized Models}\label{app:training_details:spl_bonito}
To train the task-specialized Mistral-7B-Instruct-v0.2$_{\mathrm{special}}$ and Mistral-7B$_{\mathrm{special}}$, we create a task-specific dataset by filtering out task types from the CTGA dataset. 
We selected datasets containing templates that correspond to three task types: yes-no question answering, extractive question answering, and natural language inference.
The datasets have a total of 130,703 examples for yes-no question answering, 378,167 examples for extractive question answering, and 100,250 examples for natural language inference.

To train the task-specialized \sys$_{\mathrm{special}}$, we convert the same task templates into meta templates.
Then, we use the meta templates to generate the dataset to train the model. 

For fairness, we use the same hyperparameters to train task-specialized Bonito and the task-specialized Mistral-7B-Instruct-v0.2$_\mathrm{special}$ and Mistral-7B$_\mathrm{special}$ models.
Since the datasets have significantly fewer examples than CTGA, we train these models for at most 10,000 steps.
If the training mixture has less than 160,000 examples, we train the \sys model for 1 epoch.
The training on four GPUs takes about 4 to 10 hours. 
For the rest of the hyperparameters, see Appendix \ref{app:training_details:hyperparameters}.

\subsection{Software and Hardware Details}
Our codebase is built using the transformers~\citep{wolf:arxiv19} library in PyTorch~\citep{paszke:neurips19}.
We train all the models in a distributed multi-GPU environment using DeepSpeed~\citep{rasley:kdd20}.
We use the distributed data parallel in DeepSpeed to increase the effective batch size during training. 
For training and evaluation, we use the following GPUs depending on their availability on our compute cluster: NVIDIA GeForce RTX 3090, NVIDIA RTX A5500, NVIDIA RTX A6000, NVIDIA RTX A5000, and NVIDIA A40.

\subsection{Hyperparameters}\label{app:training_details:hyperparameters}
Throughout our fine-tuning experiments, unless otherwise mentioned, we use the hyperparameters from ~\citet{dettmers:arxiv23}.
Table \ref{tab:training_details:hyperparameters} shows the hyperparameters in our experiments.
We use gradient accumulation to achieve the effective batch size of 16. 
We also use gradient checkpointing to train large models like Llama 2 7B and Mistral-7B.

%% file: tables/hyperparameters.tex
\begin{table}[]
    \centering
    \begin{tabular}{lr}\toprule
        \textbf{Hyperparameters} & \textbf{Values} \\\midrule
        Q-LoRA rank (r) & 64 \\
        Q-LoRA scaling factor ($\alpha$) & 4 \\
        Q-LoRA dropout & 0 \\
        Optimizer & Paged AdamW \\
        Learning rate scheduler & linear \\
        Max. learning rate & $1e-04$ \\
        Min. learning rate & 0 \\
        Weight decay & 0 \\
        Dropout & 0 \\
        Max. gradient norm & 0.3 \\
        Effective batch size & 16 \\
        Max. input length & 2,048 \\
        Max. output length & 2,048 \\\bottomrule
    \end{tabular}
    \caption{The hyperparameters used to train all the models in our experiments.}
    \label{tab:training_details:hyperparameters}
\end{table}

%% file: tables/task_distribution.tex
\begin{table}[h]
    \centering
    \resizebox{1.0\linewidth}{!}{
    \begin{tabular}{lr}\toprule
    \textbf{Task type} & \textbf{\# Examples} \\\midrule
    Summarization & 284,589\\
    Sentiment & 233,530\\
    Multiple-choice question answering & 229,066\\
    Extractive question answering & 222,769\\
    Topic classification & 209,980\\
    Natural language inference & 100,250\\
    Question generation & 92,847\\
    Text generation & 86,835\\
    Question answering without choices & 75,159\\
    Paraphrase identification & 47,848\\
    Sentence completion & 30,246\\
    Yes-no question answering & 25,895\\
    Word sense disambiguation & 5,428\\
    Paraphrase generation & 2,550\\
    Textual entailment & 2,490\\
    Coreference resolution & 554\\\midrule
    Total & 1,650,036\\\bottomrule
    \end{tabular}
    }
    \caption{Task distribution in the conditional task generation with attributes dataset.}
    \label{tab:task_distribution}
\end{table}

%% file: appendix/use_of_ai.tex
\section{Use of AI Assistants}
Our work used AI Assistants such as ChatGPT and Grammarly for spell-checking and fixing minor grammatical mistakes. 
We also use GitHub Co-Pilot in VSCode to write our codebase. 

%% file: appendix/ctga.tex
\section{Conditional Task Generation with Attributes: Datasets and Tasks}\label{app:ctga}
Table \ref{tab:task_distribution} shows the task distribution of the conditional task generation with attributes dataset. 
Table \ref{tab:tasks_one} lists all the datasets along with the task types in the dataset. 
The dataset includes 16 task types across 38 datasets. 
The task types are summarization, sentiment analysis, multiple-choice question answering, extractive question answering, topic classification, natural language inference, question generation, text generation, question answering without choices, paraphrase identification, sentence completion, yes-no question answering, word sense disambiguation, paraphrase generation, textual entailment, and
coreference resolution. 
The difference between extractive question answering and question answering without choices is that in extractive question answering the target answer is present in the context whereas in question answering without choices, that always is not the case. 

\begin{table*}[t]
    \centering
    \begin{tabular}{ll}\toprule
    \textbf{Dataset name} & \textbf{Task types} \\\midrule
adversarial\_qa/dbert & Extractive question answering\\
 & Question generation\\\midrule
adversarial\_qa/dbidaf & Extractive question answering\\
 & Question generation\\\midrule
adversarial\_qa/droberta & Extractive question answering\\
 & Question generation\\\midrule
ag\_news & Topic classification\\\midrule
amazon\_polarity & Sentiment\\\midrule
anli & Natural language inference\\\midrule
app\_reviews & Multiple-choice question answering\\
 & Question answering without choices\\
 & Text generation\\\midrule
cnn\_dailymail/3.0.0 & Summarization\\
 & Text generation\\\midrule
cosmos\_qa & Multiple-choice question answering\\
 & Question answering without choices\\
 & Question generation\\\midrule
dbpedia\_14 & Topic classification\\\midrule
dream & Multiple-choice question answering\\
 & Text generation\\\midrule
duorc/ParaphraseRC & Extractive question answering\\
 & Question generation\\
 & Summarization\\
 & Text generation\\\midrule
duorc/SelfRC & Extractive question answering\\
 & Question generation\\
 & Summarization\\
 & Text generation\\\midrule
gigaword & Summarization\\
 & Text generation\\\midrule
glue/mrpc & Paraphrase generation\\
 & Paraphrase identification\\\midrule
hellaswag & Sentence completion\\
 & Topic classification\\\midrule
imdb & Sentiment\\\midrule
multi\_newspaws/labeled\_final & Paraphrase generation\\
 & Paraphrase identification\\\midrule
qasc & Multiple-choice question answering\\\midrule
    \end{tabular}
    \caption{Dataset names and the prompted task types in the dataset [1/2].}
    \label{tab:tasks_one}
\end{table*}

\begin{table*}[t]
    \centering
    \begin{tabular}{ll}\toprule
    \textbf{Dataset name} & \textbf{Task types} \\\midrule
quail & Multiple-choice question answering\\
 & Question answering without choices\\\midrule
quartz & Multiple-choice question answering\\\midrule
quoref & Extractive question answering\\
 & Summarization\\\midrule
race/all & Multiple-choice question answering\\
 & Question answering without choices\\
 & Question generation\\
 & Yes-no question answering\\\midrule
ropes & Extractive question answering\\\midrule
rotten\_tomatoes & Sentiment\\\midrule
samsum & Summarization\\
 & Text generation\\\midrule
social\_i\_qa & Multiple-choice question answering\\
 & Question answering without choices\\
 & Question generation\\
 & Yes-no question answering\\\midrule
squad & Extractive question answering\\
 & Question generation\\\midrule
super\_glue/boolq & Yes-no question answering\\\midrule
super\_glue/cb & Natural language inference\\\midrule
super\_glue/copa & Sentence completion\\\midrule
super\_glue/record & Extractive question answering\\
 & Multiple-choice question answering\\\midrule
super\_glue/rte & Textual entailment\\\midrule
super\_glue/wic & Word sense disambiguation\\\midrule
super\_glue/wsc.fixed & Coreference resolution\\\midrule
wiki\_hop/original & Multiple-choice question answering\\
 & Question answering without choices\\\midrule
xsum & Summarization\\\midrule
yelp\_review\_full & Sentiment\\\midrule
    \end{tabular}
    \caption{Dataset names and the prompted task types in the dataset [2/2].}
    \label{tab:tasks_two}
\end{table*}

%% file: appendix/target_prompts.tex
\section{Prompts for Evaluation}\label{app:prompt_eval}

\subsection{{PubmedQA}}

Dataset from \citet{jin:emnlp19}:

\begin{itemize}[leftmargin=*]

\item Input 
\begin{minted}[breaklines, tabsize=2,breaksymbolleft=, fontsize=\small,bgcolor=bg]{django}
Given a passage: {{ context.contexts | join(" ") }}

Answer the question: {{question}}

Summarize the above answer as YES, NO, or MAYBE? 
\end{minted}

Target
\begin{minted}[breaklines, tabsize=2,breaksymbolleft=, fontsize=\small,bgcolor=bg]{django}
{{final_decision}}
\end{minted}
 
Answer Choices
\begin{minted}[breaklines, tabsize=2,breaksymbolleft=, fontsize=\small, bgcolor=bg]{django}
yes ||| no ||| maybe
\end{minted}

\item Input
\begin{minted}[breaklines, tabsize=2,breaksymbolleft=, fontsize=\small,bgcolor=bg]{django}
I'm a doctor and I want to answer the question "{{question}}" using The following passage:

{{ context.contexts | join(" ") }}

Summarize the above answer as YES, NO, or MAYBE? 
\end{minted}

Target 
\begin{minted}[breaklines, tabsize=2,breaksymbolleft=, fontsize=\small,bgcolor=bg]{django}
{{final_decision}}
\end{minted}

Answer Choices
\begin{minted}[breaklines, tabsize=2,breaksymbolleft=, fontsize=\small, bgcolor=bg]{django}
yes ||| no ||| maybe
\end{minted}

\item Input 
\begin{minted}[breaklines, tabsize=2,breaksymbolleft=, fontsize=\small,bgcolor=bg]{django}
What is the answer to the question "{{question}}" based on The following passage:

{{ context.contexts | join(" ") }}

Summarize the above answer as YES, NO, or MAYBE?
\end{minted}

Target 
\begin{minted}[breaklines, tabsize=2,breaksymbolleft=, fontsize=\small,bgcolor=bg]{django}
{{final_decision}}
\end{minted}

Answer Choices 
\begin{minted}[breaklines, tabsize=2,breaksymbolleft=, fontsize=\small, bgcolor=bg]{django}
yes ||| no ||| maybe
\end{minted}

\item Input 
\begin{minted}[breaklines, tabsize=2,breaksymbolleft=, fontsize=\small,bgcolor=bg]{django}
Please answer the question "{{question}}" using The following passage:

{{ context.contexts | join(" ") }}

Summarize the above answer as YES, NO, or MAYBE?
\end{minted}

Target 
\begin{minted}[breaklines, tabsize=2,breaksymbolleft=, fontsize=\small,bgcolor=bg]{django}
{{final_decision}}
\end{minted}

Answer Choices
\begin{minted}[breaklines, tabsize=2,breaksymbolleft=, fontsize=\small, bgcolor=bg]{django}
yes ||| no ||| maybe
\end{minted}

\item Input 
\begin{minted}[breaklines, tabsize=2,breaksymbolleft=, fontsize=\small,bgcolor=bg]{django}
Given the following passage, answer the question: "{{question}}"

            Passage: {{ context.contexts | join(" ") }}

Summarize the above answer as YES, NO, or MAYBE?
\end{minted}

Target 
\begin{minted}[breaklines, tabsize=2,breaksymbolleft=, fontsize=\small,bgcolor=bg]{django}
{{final_decision}}
\end{minted}

Answer Choices
\begin{minted}[breaklines, tabsize=2,breaksymbolleft=, fontsize=\small, bgcolor=bg]{django}
yes ||| no ||| maybe
\end{minted}

\end{itemize}

\subsection{{Privacy Policy QA}}

Dataset from \citet{ravichander:emnlp19}.

\begin{itemize}[leftmargin=*]

\item Input 
\begin{minted}[breaklines, tabsize=2,breaksymbolleft=, fontsize=\small,bgcolor=bg]{django}
Given the context, is this related to the question?
Context: {{text}}
Question: {{question}}
\end{minted}

Target 
\begin{minted}[breaklines, tabsize=2,breaksymbolleft=, fontsize=\small,bgcolor=bg]{django}
{{answer}}
\end{minted}

Answer Choices
\begin{minted}[breaklines, tabsize=2,breaksymbolleft=, fontsize=\small, bgcolor=bg]{django}
Relevant|||Irrelevant
\end{minted}

\item Input 
\begin{minted}[breaklines, tabsize=2,breaksymbolleft=, fontsize=\small,bgcolor=bg]{django}
Is this question
"{{question}}"
related to this context
"{{text}}"?
\end{minted}

Target 
\begin{minted}
[breaklines, tabsize=2,breaksymbolleft=, fontsize=\small,bgcolor=bg]{django}
{%
\end{minted}

Answer Choices 
\begin{minted}[breaklines, tabsize=2,breaksymbolleft=, fontsize=\small, bgcolor=bg]{django}
Yes|||No
\end{minted}

\item Input 
\begin{minted}[breaklines, tabsize=2,breaksymbolleft=, fontsize=\small,bgcolor=bg]{django}
Can this
"{{text}}"
help answer this question
"{{question}}"?
\end{minted}

Target
\begin{minted}[breaklines, tabsize=2,breaksymbolleft=, fontsize=\small,bgcolor=bg]{django}
{%
\end{minted}

Answer Choices
\begin{minted}[breaklines, tabsize=2,breaksymbolleft=, fontsize=\small, bgcolor=bg]{django}
Yes|||No
\end{minted}

\item Input 
\begin{minted}[breaklines, tabsize=2,breaksymbolleft=, fontsize=\small,bgcolor=bg]{django}
As a lawyer, can you answer the question given the context?
Question: {{question}}
Context:{{text}}
\end{minted}

Target 
\begin{minted}[breaklines, tabsize=2,breaksymbolleft=, fontsize=\small,bgcolor=bg]{django}
{%
\end{minted}

Answer Choices 
\begin{minted}[breaklines, tabsize=2,breaksymbolleft=, fontsize=\small, bgcolor=bg]{django}
Yes|||No
\end{minted}

\item Input 
\begin{minted}[breaklines, tabsize=2,breaksymbolleft=, fontsize=\small,bgcolor=bg]{django}
Question:{{question}}
Context:{{text}}
Is the question related to the context?
\end{minted}

Target
\begin{minted}[breaklines, tabsize=2,breaksymbolleft=, fontsize=\small,bgcolor=bg]{django}
{%
\end{minted}

Answer Choices
\begin{minted}[breaklines, tabsize=2,breaksymbolleft=, fontsize=\small, bgcolor=bg]{django}
Yes|||No
\end{minted}

\end{itemize}

\subsection{{SQuADShifts}}

Dataset from \citet{miller:icml20}. 

\subsubsection{{NYT}}
\begin{itemize}[leftmargin=*]

\item Input 
\begin{minted}[breaklines, tabsize=2,breaksymbolleft=, fontsize=\small,bgcolor=bg]{django}
After reading the following paragraph, please answer this question: {{question}}

{{context}}
\end{minted}

Target 
\begin{minted}[breaklines, tabsize=2,breaksymbolleft=, fontsize=\small,bgcolor=bg]{django}
{{answers['text'] | most_frequent | choice}}
\end{minted}

\item Input 
\begin{minted}[breaklines, tabsize=2,breaksymbolleft=, fontsize=\small,bgcolor=bg]{django}
I'm working on the final exam for my class and am trying to figure out the answer to the question "{{question}}" I found the following info on New York Times and I think it has the answer. Can you tell me the answer?

{{context}}

\end{minted}

Target
\begin{minted}[breaklines, tabsize=2,breaksymbolleft=, fontsize=\small,bgcolor=bg]{django}
{{answers['text'] | most_frequent | choice}}
\end{minted}

\item Input 
\begin{minted}[breaklines, tabsize=2,breaksymbolleft=, fontsize=\small,bgcolor=bg]{django}
I've always wondered: {{question}}

I searched New York Times and this is what I found. What's the answer?

{{context}}

\end{minted}

Target
\begin{minted}[breaklines, tabsize=2,breaksymbolleft=, fontsize=\small,bgcolor=bg]{django}
{{answers['text'] | most_frequent | choice}}
\end{minted}

\item Input 
\begin{minted}[breaklines, tabsize=2,breaksymbolleft=, fontsize=\small,bgcolor=bg]{django}
{{context}}

With the help of the passage, please answer the following question: 
{{question}} 
\end{minted}

Target
\begin{minted}[breaklines, tabsize=2,breaksymbolleft=, fontsize=\small,bgcolor=bg]{django}
{{answers["text"]|choice}}

\end{minted}

\item Input 
\begin{minted}[breaklines, tabsize=2,breaksymbolleft=, fontsize=\small,bgcolor=bg]{django}
{{["Question", "Problem"]  | choice}} {{range(1, 12) | choice}}: {{question}}

Hint: {{context}}

\end{minted}

Target 
\begin{minted}[breaklines, tabsize=2,breaksymbolleft=, fontsize=\small,bgcolor=bg]{django}
{{answers["text"] | most_frequent | choice}}
\end{minted}

\end{itemize}

\subsubsection{{Amazon}}

\begin{itemize}[leftmargin=*]

\item Input 
\begin{minted}[breaklines, tabsize=2,breaksymbolleft=, fontsize=\small,bgcolor=bg]{django}
After reading the following paragraph, please answer this question: {{question}}

{{context}}

\end{minted}

Target 
\begin{minted}[breaklines, tabsize=2,breaksymbolleft=, fontsize=\small,bgcolor=bg]{django}
{{answers['text'] | most_frequent | choice}}
\end{minted}

\item Input 
\begin{minted}[breaklines, tabsize=2,breaksymbolleft=, fontsize=\small,bgcolor=bg]{django}
I'm working on the final exam for my class and am trying to figure out the answer to the question "{{question}}" I found the following info on Amazon and I think it has the answer. Can you tell me the answer?

{{context}}

\end{minted}

Target
\begin{minted}[breaklines, tabsize=2,breaksymbolleft=, fontsize=\small,bgcolor=bg]{django}
{{answers['text'] | most_frequent | choice}}
\end{minted}

\item Input 
\begin{minted}[breaklines, tabsize=2,breaksymbolleft=, fontsize=\small,bgcolor=bg]{django}
I've always wondered: {{question}}

I searched Amazon and this is what I found. What's the answer?

{{context}}

\end{minted}

Target 
\begin{minted}[breaklines, tabsize=2,breaksymbolleft=, fontsize=\small,bgcolor=bg]{django}
{{answers['text'] | most_frequent | choice}}
\end{minted}

\item Input 
\begin{minted}[breaklines, tabsize=2,breaksymbolleft=, fontsize=\small,bgcolor=bg]{django}
{{context}}

With the help of the passage, please answer the following question: 
{{question}} 
\end{minted}

Target 
\begin{minted}[breaklines, tabsize=2,breaksymbolleft=, fontsize=\small,bgcolor=bg]{django}
{{answers["text"]|choice}}

\end{minted}

\item Input 
\begin{minted}[breaklines, tabsize=2,breaksymbolleft=, fontsize=\small,bgcolor=bg]{django}
{{["Question", "Problem"]  | choice}} {{range(1, 12) | choice}}: {{question}}

Hint: {{context}}

\end{minted}

Target 
\begin{minted}[breaklines, tabsize=2,breaksymbolleft=, fontsize=\small,bgcolor=bg]{django}
{{answers["text"] | most_frequent | choice}}
\end{minted}

\end{itemize}

\subsubsection{{Reddit}}

\begin{itemize}[leftmargin=*]

\item  Input
\begin{minted}[breaklines, tabsize=2,breaksymbolleft=, fontsize=\small,bgcolor=bg]{django}
After reading the following paragraph, please answer this question: {{question}}

{{context}}

\end{minted}

Target
\begin{minted}[breaklines, tabsize=2,breaksymbolleft=, fontsize=\small,bgcolor=bg]{django}
{{answers['text'] | most_frequent | choice}}
\end{minted}

\item  Input
\begin{minted}[breaklines, tabsize=2,breaksymbolleft=, fontsize=\small,bgcolor=bg]{django}
I'm working on the final exam for my class and am trying to figure out the answer to the question "{{question}}" I found the following info on Reddit and I think it has the answer. Can you tell me the answer?

{{context}}

\end{minted}

Target
\begin{minted}[breaklines, tabsize=2,breaksymbolleft=, fontsize=\small,bgcolor=bg]{django}
{{answers['text'] | most_frequent | choice}}
\end{minted}

\item  Input
\begin{minted}[breaklines, tabsize=2,breaksymbolleft=, fontsize=\small,bgcolor=bg]{django}
I've always wondered: {{question}}

I searched Reddit and this is what I found. What's the answer?

{{context}}

\end{minted}

Target
\begin{minted}[breaklines, tabsize=2,breaksymbolleft=, fontsize=\small,bgcolor=bg]{django}
{{answers['text'] | most_frequent | choice}}
\end{minted}

\item  Input
\begin{minted}[breaklines, tabsize=2,breaksymbolleft=, fontsize=\small,bgcolor=bg]{django}
{{context}}

With the help of the passage, please answer the following question: 
{{question}} 
\end{minted}

Target
\begin{minted}[breaklines, tabsize=2,breaksymbolleft=, fontsize=\small,bgcolor=bg]{django}
{{answers["text"]|choice}}

\end{minted}

\item Input
\begin{minted}[breaklines, tabsize=2,breaksymbolleft=, fontsize=\small,bgcolor=bg]{django}
{{["Question", "Problem"]  | choice}} {{range(1, 12) | choice}}: {{question}}

Hint: {{context}}

\end{minted}

Target 
\begin{minted}[breaklines, tabsize=2,breaksymbolleft=, fontsize=\small,bgcolor=bg]{django}
{{answers["text"] | most_frequent | choice}}
\end{minted}

\end{itemize}

\subsection{{ContractNLI}}

Dataset from \citet{koreeda:emnlp21}.

\begin{itemize}[leftmargin=*]
\item Input 
\begin{minted}[breaklines, tabsize=2,breaksymbolleft=, fontsize=\small,bgcolor=bg]{django}
Suppose {{premise}} Can we infer that "{{hypothesis}}"? yes, no or maybe?
\end{minted}

Target 
\begin{minted}[breaklines, tabsize=2,breaksymbolleft=, fontsize=\small,bgcolor=bg]{django}
{{answer_choices[label]}}
\end{minted}

Answer Choices 
\begin{minted}[breaklines, tabsize=2,breaksymbolleft=, fontsize=\small, bgcolor=bg]{django}
No ||| Yes ||| Maybe
\end{minted}

\item Input 
\begin{minted}[breaklines, tabsize=2,breaksymbolleft=, fontsize=\small,bgcolor=bg]{django}
{{premise}} 

Question: Does this imply that "{{hypothesis}}"? yes, no or maybe?
\end{minted}

Target
\begin{minted}[breaklines, tabsize=2,breaksymbolleft=, fontsize=\small,bgcolor=bg]{django}
{{answer_choices[label]}}
\end{minted}

Answer Choices
\begin{minted}[breaklines, tabsize=2,breaksymbolleft=, fontsize=\small, bgcolor=bg]{django}
No ||| Yes ||| Maybe
\end{minted}

\item Input 
\begin{minted}[breaklines, tabsize=2,breaksymbolleft=, fontsize=\small,bgcolor=bg]{django}
Take the following as truth: {{premise}} Then the following statement: "{{hypothesis}}" is {{"true"}}, {{"false"}}, or {{"inconclusive"}}?
\end{minted}

Target 
\begin{minted}[breaklines, tabsize=2,breaksymbolleft=, fontsize=\small,bgcolor=bg]{django}
{{answer_choices[label]}}
\end{minted}

Answer Choices 
\begin{minted}[breaklines, tabsize=2,breaksymbolleft=, fontsize=\small, bgcolor=bg]{django}
False ||| True ||| Inconclusive
\end{minted}

\item Input 
\begin{minted}[breaklines, tabsize=2,breaksymbolleft=, fontsize=\small,bgcolor=bg]{django}
{{premise}} Based on that information, is the claim: "{{hypothesis}}" {{"true"}}, {{"false"}}, or {{"inconclusive"}}? 
\end{minted}

Target 
\begin{minted}[breaklines, tabsize=2,breaksymbolleft=, fontsize=\small,bgcolor=bg]{django}
{{ answer_choices[label]}}
\end{minted}

Answer Choices 
\begin{minted}[breaklines, tabsize=2,breaksymbolleft=, fontsize=\small, bgcolor=bg]{django}
False ||| True ||| Inconclusive
\end{minted}

\item Input 
\begin{minted}[breaklines, tabsize=2,breaksymbolleft=, fontsize=\small,bgcolor=bg]{django}
{{premise}} Based on the previous passage, is it true that "{{hypothesis}}"? Yes, no, or maybe? 
\end{minted}

Target 
\begin{minted}[breaklines, tabsize=2,breaksymbolleft=, fontsize=\small,bgcolor=bg]{django}
 {{ answer_choices[label] }}
\end{minted}

Answer Choices 
\begin{minted}[breaklines, tabsize=2,breaksymbolleft=, fontsize=\small, bgcolor=bg]{django}
No ||| Yes ||| Maybe
\end{minted}

\end{itemize}

\subsection{{Vitamin C}}

Dataset from \citet{schuster:acl21}. 

\begin{itemize}[leftmargin=*]

\item Input 
\begin{minted}[breaklines, tabsize=2,breaksymbolleft=, fontsize=\small,bgcolor=bg]{django}
Suppose {{evidence}} Can we infer that "{{claim}}"? yes, no or maybe?
\end{minted}

Target 
\begin{minted}[breaklines, tabsize=2,breaksymbolleft=, fontsize=\small,bgcolor=bg]{django}
{%
\end{minted}

Answer Choices
\begin{minted}[breaklines, tabsize=2,breaksymbolleft=, fontsize=\small, bgcolor=bg]{django}
No ||| Yes ||| Maybe
\end{minted}

\item Input 
\begin{minted}[breaklines, tabsize=2,breaksymbolleft=, fontsize=\small,bgcolor=bg]{django}
{{evidence}} 

Question: Does this imply that "{{claim}}"? yes, no or maybe?
\end{minted}

Target
\begin{minted}[breaklines, tabsize=2,breaksymbolleft=, fontsize=\small,bgcolor=bg]{django}
{%
\end{minted}

Answer Choices
\begin{minted}[breaklines, tabsize=2,breaksymbolleft=, fontsize=\small, bgcolor=bg]{django}
No ||| Yes ||| Maybe
\end{minted}

\item Input 
\begin{minted}[breaklines, tabsize=2,breaksymbolleft=, fontsize=\small,bgcolor=bg]{django}
Take the following as truth: {{evidence}} Then the following statement: "{{claim}}" is {{"true"}}, {{"false"}}, or {{"inconclusive"}}?
\end{minted}

Target 
\begin{minted}[breaklines, tabsize=2,breaksymbolleft=, fontsize=\small,bgcolor=bg]{django}
{%
\end{minted}

Answer Choices
\begin{minted}[breaklines, tabsize=2,breaksymbolleft=, fontsize=\small, bgcolor=bg]{django}
False ||| True ||| Inconclusive
\end{minted}

\item Input 
\begin{minted}[breaklines, tabsize=2,breaksymbolleft=, fontsize=\small,bgcolor=bg]{django}
{{evidence}}
Based on that information, is the claim: "{{claim}}" {{"true"}}, {{"false"}}, or {{"inconclusive"}}? 
\end{minted}

Target 
\begin{minted}[breaklines, tabsize=2,breaksymbolleft=, fontsize=\small,bgcolor=bg]{django}
{%
\end{minted}

Answer Choices
\begin{minted}[breaklines, tabsize=2,breaksymbolleft=, fontsize=\small, bgcolor=bg]{django}
False ||| True ||| Inconclusive
\end{minted}

\item Input 
\begin{minted}[breaklines, tabsize=2,breaksymbolleft=, fontsize=\small,bgcolor=bg]{django}
{{evidence}} Based on the previous passage, is it true that "{{claim}}"? Yes, no, or maybe? 
\end{minted}

Target
\begin{minted}[breaklines, tabsize=2,breaksymbolleft=, fontsize=\small,bgcolor=bg]{django}
{%
\end{minted}

Answer Choices
\begin{minted}[breaklines, tabsize=2,breaksymbolleft=, fontsize=\small, bgcolor=bg]{django}
No ||| Yes ||| Maybe
\end{minted}

\end{itemize}

%% file: appendix/qualitative.tex
\section{Qualitatitve Examples}\label{app:qualitative}
Table \ref{tab:qualitative} shows \sys-generated tasks for the PubMedQA, SQuADShifts Amazon, and ContractNLI.

\input{tables/qualitative}

%% file: tables/qualitative.tex
\begin{table*}[t!]
\fbox{
\begin{minipage}{\linewidth}
\begin{small}
\begin{lmttfont}
\underline{\textbf{Dataset:}} PubMedQA\\
\underline{\textbf{Task type:}} Yes-no Question Answering \\
\underline{\textbf{Input:}}
Palmitate, a saturated fatty acid (FA), is known to induce toxicity and cell death in various types of cells. Resveratrol (RSV) is able to prevent pathogenesis and/or decelerate the progression of a variety of diseases. Several in vitro and in vivo studies have also shown a protective effect of RSV on fat accumulation induced by FAs. Additionally, endoplasmic reticulum (ER) stress has recently been linked to cellular adipogenic responses. To address the hypothesis that the RSV effect on excessive fat accumulation promoted by elevated saturated FAs could be partially mediated by a reduction of ER stress, we studied the RSV action on experimentally induced ER stress using palmitate in several cancer cell lines. We show that, unexpectedly, RSV promotes an amplification of palmitate toxicity and cell death and that this mechanism is likely due to a perturbation of palmitate accumulation in the triglyceride form and to a less important membrane fluidity variation. Additionally, RSV decreases radical oxygen species (ROS) generation in palmitate-treated cells but leads to enhanced X-box binding protein-1 (XBP1) splicing and C/EBP homologous protein (CHOP) expression. These molecular effects are induced simultaneously to caspase-3 cleavage, suggesting that RSV promotes palmitate lipoapoptosis primarily through an ER stress-dependent mechanism. Moreover, the lipotoxicity reversion induced by eicosapentaenoic acid (EPA) or by a liver X receptor (LXR) agonist reinforces the hypothesis that RSV-mediated inhibition of palmitate channeling into triglyceride pools could be a key factor in the aggravation of palmitate-induced cytotoxicity.Our results suggest that RSV exerts its cytotoxic role in cancer cells exposed to a saturated FA context primarily by triglyceride accumulation inhibition, probably leading to an intracellular palmitate accumulation that triggers a lipid-mediated cell death. Additionally, this cell death is promoted by ER stress through a CHOP-mediated apoptotic process and may represent a potential anticancer strategy.\\
Question: does resveratrol induce cell death in cancer cells\\
Answer:\\
\underline{\textbf{Output:}} Yes
\end{lmttfont}
\end{small}
\end{minipage}
}
\fbox{
\begin{minipage}{\linewidth}
\begin{small}
\begin{lmttfont}
\underline{\textbf{Dataset:}} SquadShifts Amazon\\
\underline{\textbf{Task type:}} Extractive question answering \\
\underline{\textbf{Input:}} Here is a review left by a customer on a product. Would you be able to answer the question asked by the customer about the product?\\
Review: The storage bag I got is different from the picture. It's a mesh see-through material, where in the picture it doesn't look like it would be see-through. It still does what it's supposed to, so no complaints. I love the velcro loop at the top that I can use to hang it on the bar under the shelf in my closet.\\
Question: How did the customer hang the storage bag?\\
\underline{\textbf{Output:}} on the bar under the shelf
\end{lmttfont}
\end{small}
\end{minipage}
}
\fbox{
\begin{minipage}{\linewidth}
\begin{small}
\begin{lmttfont}
\underline{\textbf{Dataset:}} Contract NLI\\
\underline{\textbf{Task type:}} Natural language infernce\\
\underline{\textbf{Input:}} 4. Nothing in this Agreement is to be construed as granting the Recipient, by implication or otherwise, any right whatsoever with respect to the Confidential Information or part thereof. Using only the above description and what you know about the world, "The recipient is a person." is definitely correct, incorrect, or inconclusive?\\
\underline{\textbf{Output:}} Inconclusive
\end{lmttfont}
\end{small}
\end{minipage}
}
\caption{
Example generations from \sys for PubMedQA, SQuADShifts Amazon, and ContractNLI. 
}
\label{tab:qualitative}
\end{table*}